\documentclass[journal]{IEEEtran}

\ifCLASSINFOpdf
\else
\fi

\usepackage{acro}
\usepackage[utf8]{inputenc}
\usepackage[ruled,vlined,linesnumbered]{algorithm2e}
\usepackage{amsfonts, amsmath, amssymb}
\usepackage{array}
\usepackage{bm}
\usepackage{cite}
\usepackage{color}
\usepackage{float}
\usepackage{graphicx}
\usepackage{subfig}
\usepackage{tablefootnote}
\usepackage{hyperref}
\usepackage{url}
\usepackage{multirow}
\usepackage{booktabs}
\usepackage{tabularx}
\usepackage{flushend} 
\usepackage{threeparttable}

\usepackage[export]{adjustbox}

\definecolor{purple}{RGB}{106,13,173}

\usepackage{graphicx}

\graphicspath{{./Images/}}

\begin{document}

\title{snnTrans-DHZ: A Lightweight Spiking Neural Network Architecture for Underwater Image Dehazing}

\author{
    Vidya Sudevan$^{1*}$, Fakhreddine Zayer$^{1}$, Rizwana Kausar$^{1}$, Sajid Javed$^{1}$, Hamad Karki$^{1}$,\\ Giulia De Masi$^{2}$, Jorge Dias$^{1}$ \\
     $^{1}$ Center for Autonomous Robotic Systems, Khalifa University, Abu Dhabi, UAE\\
     $^{2}$ Dept. Science and Engineering, Sorbonne University, Abu Dhabi, UAE\\
}


\maketitle

\begin{abstract}
Underwater image dehazing is critical for any vision-based marine environment, where light scattering and absorption significantly degrade visibility. This paper presents snnTrans-DHZ, a lightweight Spiking Neural Network (SNN) architecture specifically designed for underwater image dehazing. Using the temporal dynamics of SNNs, snnTrans-DHZ efficiently processes time-dependent raw image sequences while maintaining low power consumption. The raw underwater images are first converted into time-dependent image sequences by repeatedly passing the static image to a user-defined timestep value. The RGB sequences are then converted into LAB color space representations and processed simultaneously. The architecture integrates three primary modules:(i) \textit{K estimator} to extract features from different color space representations,(ii) \textit{Background Light Estimator} To jointly estimate the background light component from the RGB-LAB color space representations, and (iii) \textit{soft image reconstruction} to reconstruct the haze-free, visibility-enhanced image. snnTrans-DHZ model is directly trained using surrogate gradient-based backpropagation through time (BPTT) strategy. In this research, a combined loss function is designed and used. Our model is trained and tested on the UIEB, a publicly available benchmark dataset. The snnTrans-DHZ algorithm achieves PSNR = \(21.6773~\text{dB}\) and SSIM = \(0.8795\) on UIEB, and PSNR = \(23.4562~\text{dB}\) and SSIM = \(0.8439\) on EUVP. The total network parameters in the snnTrans-DHZ architecture are \textbf{\(0.5670~M\)}. Moreover, snnTrans-DHZ achieves this algorithmic performance with fewer operations (\(7.42\) GSOPs) and lower energy consumption (\(0.0151~J\)) compared to existing state-of-the-art image enhancement methods. Compared to conventional learning-based dehazing methods that use CNNs and vision transformers, the lightweight design and low power consumption of the SNN-DHZ model make it ideal for deployment in underwater robotics, marine exploration, and environmental monitoring. 
\end{abstract}


\IEEEpeerreviewmaketitle

\section{Introduction}

Underwater image dehazing remains a critical challenge in computer vision due to poor lighting, color distortion, and water turbidity, all of which degrade image quality \cite{zhou2023underwater}. Methods like Shallow-UWnet \cite{naik2021shallow}, WaterGAN \cite{li2017watergan}, U-Trans \cite{peng2023u}, and WaterFormer \cite{wu2024underwater} have introduced novel architectures, including shallow ConvBlocks, probabilistic learning, GAN-based pipelines, and transformer-based modules. Although these approaches are effective, they often fall short in energy efficiency, necessitating the exploration of novel computational paradigms. 

Spiking Neural Networks (SNNs) are energy-efficient systems for spatio-temporal processing, leveraging temporal spike timing for reduced energy consumption \cite{roy2019towards}. Efficient SNNs for image processing are developed via ANN-to-SNN conversion, using techniques like weight normalization, or direct training with surrogate gradients \cite{li2024deep}. While recent work has shown promise in image reconstruction and denoising, these methods are limited to low-resolution images and Gaussian noise removal \cite{castagnetti2023spiden}. To address these limitations, this study explores SNN-based approaches for high-resolution underwater image dehazing.

The key contributions include:
\begin{enumerate}
    \item \textbf{Theoretical Advancements:}
    \begin{itemize}
    \item \textit{Adaptive Leaky Integrate-and-Fire (ALIF) Neurons:} Integration of ALIF neurons with a learnable threshold membrane potential enhances temporal dynamics and memory retention within the snnTrans-DHZ architecture. Unlike fixed-threshold LIF neurons, ALIF neurons adapt their thresholds dynamically during backpropagation, enabling more efficient temporal modeling.
    \item \textit{Application-Specific Loss Function:} Development of a custom loss function tailored specifically for underwater image dehazing tasks, integrating perceptual quality metrics (SSIM, contrast enhancement) with traditional pixel-wise losses. This formulation ensures balanced optimization between haze removal, structural preservation, and natural color fidelity.
    \end{itemize}
    \item \textbf{Practical Implementations:}
    \begin{itemize}
    \item \textit{Lightweight SNN-based Framework (snnTrans-DHZ):} Introduction of snnTrans-DHZ, an efficient transformer-integrated Spiking Neural Network (SNN) framework for underwater image dehazing. With only \(0.5670~M\) parameters, the model demonstrates superior efficiency and performance compared to the baseline UIE-SNN framework.
    \item \textit{Hybrid RGB–LAB Processing:} Implementation of hybrid RGB-LAB color space transformation, enabling separate processing of luminance and chromaticity components. This domain-aware approach significantly improves haze removal, color correction, and overall perceptual quality in reconstructed underwater images.
    \end{itemize}
\end{enumerate}

The proposed snnTrans-DHZ algorithm addresses key limitations in existing transformer-based \cite{sun2024ghost, kulkarni2024multi,gao2024ddformer} image dehazing methods by operating fully within the spiking domain, significantly enhancing energy and parameter efficiency. Unlike conventional transformer architectures that process continuous-valued signals through Multiply-And-Accumulate (MAC) operations, snnTrans-DHZ utilizes binary spikes (0 or 1 valued feature maps), thus replacing computationally intensive MAC operations with simpler accumulation operations. Adaptive Leaky Integrate-and-Fire (ALIF) neurons are strategically integrated after each layer within the multi-head self-attention mechanism to maintain binary feature maps, substantially reducing the complexity associated with matrix multiplications involving real-valued feature maps. Compared specifically to the state-of-the-art UIE-SNN \cite{sudevan2025underwater} method, snnTrans-DHZ employs a shallower architecture with a depth of only two layers and utilizes significantly fewer channels (64 versus 1024), effectively minimizing parameter count without compromising performance. These innovations collectively position snnTrans-DHZ as a more computationally efficient and energy-conscious alternative for image dehazing tasks.

\section{Related Works}

\subsection{Learning-Based dehazing methods}

Li et al. \cite{li2020underwater} introduced WaterNet, a simple learning-based UIE model with only convolutional layers. To predict the confidence maps, this model processes three derived inputs along with the original input. Feature Transformation Units (FTUs) are employed to minimize color distortions and artifacts caused by white balance adjustments, histogram equalization, and gamma correction techniques. The processed inputs are then refined using confidence maps to generate the final enhanced image. Shallow-UWnet \cite{naik2021shallow} employs small convolutional blocks with Conv-ReLU-Dropout layers to enhance skip connections and is optimized using a combination of multiple loss functions, including Mean Squared Error (MSE) and Visual Geometry Group (VGG)-based perceptual loss. Fu et al. \cite{fu2022uncertainty} introduced PUIE-Net, a probabilistic model designed to analyze and enhance degraded underwater images. It incorporates a conditional variational autoencoder with adaptive instance normalization, followed by a consensus mechanism to ensure reliable output predictions. The PDCFNet \cite{zhang2024pdcfnet} method introduced Pixel Difference Convolution (PDC) to enhance high-frequency features, improving texture and detail restoration. It employs a cross-level feature fusion module to ensure effective interaction between multi-level features, leading to enhanced image quality. Dazhao et al. \cite{du2024physical} developed a physics-driven framework that simultaneously trains a "Deep Degradation Model (DDM)" alongside any UIE model. The DDM is responsible for estimating critical aspects of the underwater imaging process. It guarantees more accurate depth estimation and physically consistent improvement. The DDM and UIEConv models undergo joint training. UIEConv, a fully convolutional UIE model, enhances images by incorporating both global and local features through a dual-branch design.

With the rise of GAN models, the WaterGAN \cite{li2017watergan} model was  introduced to generate synthetic underwater images from corresponding terrestrial images  and depth information. This model follows an unsupervised learning approach to enhance color correction for monocular underwater images. It employs a depth estimation network to reconstruct relative depth maps and a color restoration network based on a fully convolutional encoder-decoder architecture, such as SegNet. PUGAN \cite{cong2023pugan} is a physical model-guided GAN architecture for UIE. It consists of a parameter estimation subnetwork that learns physical model inversion parameters by using generated color enhancement images as auxiliary information for a two-stream interaction enhancement subnetwork. Dual discriminators enforce style and content constraints, improving authenticity and visual quality. The DGD-cGAN \cite{gonzalez2024dgd} model specifically addresses underwater haze and color distortions by incorporating water-related priors in its image restoration process. It utilizes two generators: one focuses on predicting the restored scene to enhance image clarity, while the other models the underwater image formation process, employing a specialized loss function that accounts for transmission properties and veiling light effects. Finally, in LFT-DGAN \cite{zheng2024learnable} method, a learnable full-frequency domain transformer is integrated within a dual generative adversarial network. It introduces a reversible convolution-based image decomposition technique to separate underwater images into low, medium, and high-frequency domains for enhanced feature extraction. Additionally, it employs image channels and spatial similarity to facilitate cross-domain interaction and utilizes a dual-domain discriminator to learn both spatial and frequency characteristics, improving image restoration quality. 

The Ucolor \cite{li2021underwater} method learns feature representations from various color spaces and emphasizes the most discriminative features using a channel-attention module. It incorporates domain knowledge by utilizing the reverse medium transmission map as attention weights. LANet \cite{liu2022adaptive} is a supervised adaptive attention network for UIE. It begins with a multiscale fusion module that combines different spatial information. It then employs a parallel attention mechanism that focuses on illuminated features and key color variations, utilizing both pixel-wise and channel-based attention. An adaptive learning framework retains shallow details while adaptively capturing important features. The training process incorporates a multinomial loss function that combines MAE loss and perceptual loss with an asynchronous training mode to enhance the performance. The Deep-WaveNet \cite{sharma2023wavelength} model is fine-tuned with conventional pixel-wise and feature-driven cost functions. It employs a distinct receptive field structure for each image channel, driven by its wavelength, which helps in learning diverse local and global features. These features undergo refinement through a block attention mechanism. The DICAM \cite{tolie2024dicam} method addresses challenges associated with uneven degradation and color inconsistencies by utilizing inception modules on individual color channels to capture feature maps at various scales. These features are then processed through a Channel-wise Attention Module (CAM) to assess the relevance of different types of degradations. Lastly, the combined feature maps undergo further refinement with CAM to enhance the image's color richness. 

The U-Trans \cite{peng2023u} method incorporates a multiscale feature fusion transformer for channel-wise processing, coupled with a transformer module designed for global spatial feature modeling. This integration strengthens the model’s capability to handle color variations and spatial regions affected by substantial attenuation. UIE-Convformer \cite{wang2024uie} is a hybrid method combining CNNs and a feature fusion transformer. It employs a multiscale U-Net structure for extracting rich texture and semantic information, while the feature fusion transformer module handles global information fusion. A jump fusion connection module between the encoder and decoder fuses multiscale features through bidirectional cross-connection and weighted fusion, enriching the feature information for image reconstruction. The MFMN model \cite{zheng2024multi} is a multi-scale feature modulation network designed to balance model efficiency and reconstruction performance. It employs a Multi-scale Spatial Feature Module (MSFM) within a Visual Transformer-like block to dynamically extract spatial features and integrates a channel mixing module to enhance feature representation across channels. ICDT-XL/2 \cite{nie2024image} is an image-conditional diffusion transformer model for UIE tasks. It replaces the conventional U-Net in a Denoising Diffusion Probabilistic Model (DDPM) with transformer-based architecture. By converting degraded images into a latent space, it benefits from the scalability and efficiency of transformers for improved restoration. The CFPNet \cite{fu2024cfpnet} model is a complementary feature perception network that embeds a transformer into CNN-based UNet3+ to fuse local and global features effectively. It employs a dual encoder structure combining CNN and transformer branches. It includes a full-scale feature fusion module for multi-scale information merging and an auxiliary feature-guided learning strategy to enhance color correction and deblurring. The MMIETransformer \cite{kulkarni2024multi} model is a transformer-based architecture for general image enhancement, incorporating space-aware deformable convolution and multi-head self-attention for fine texture reconstruction. It consists of a spatially attentive offset extractor, an edge-enhancing feature fusion block, and a global context-aware channel attention module to refine edge details and enhance multi-stream feature learning.

The WPFNet \cite{liu2024wavelet} model, introduced by Shiben et al., utilizes a hybrid wavelet-pixel domain fusion approach to improve illumination, reduce color deviation, and improve details in degraded underwater images. It integrates a "Wavelet Domain Module (WDM)" to capture frequency-based multi-scale information, a "Residual Attention Block (RAB)" to refine low-frequency details, and a Transformer Block (T\_Block) for high-frequency feature processing. Additionally, "Pixel Domain Module (PDM)" is developed to identify and highlight detailed spatial attributes. The DDFormer \cite{gao2024ddformer} method integrates a dimension decomposition transformer with semi-supervised learning for UIE. It introduces dimension decomposition attention to compute global dependencies at the original scale and correct color distortions. A multistage transformer strategy is employed in DDFormer to capture multi-scale global information. The Ghost-Unet \cite{sun2024ghost} model introduced by Lingyu et al. is an efficient UIE framework leveraging a conditional diffusion approach to reduce redundant feature maps using linear transformations. It consists of a forward diffusion process, a reverse denoising process, and a noise prediction network, aiming to generate realistic underwater degradation images from high-quality inputs. Zhiqiang et al. proposed the Dynamic SpectraFormer \cite{hu2024dynamic} method to enhance underwater images using a frequency domain transformer combined with an ultra-high-resolution sparse spectrum attention module for long-term dependency modeling. It employs a dynamic spectrum weight generation layer that acts as an adaptive spectrum band selector, emphasizing critical frequency bands while suppressing irrelevant ones for improved enhancement. Transformer-based models demonstrate performance comparable to CNNs  by effectively capturing long-range relationships and expanding receptive areas across various visual tasks. However, their application and deployment are challenged by the large number of parameters involved.

\subsection{SNN-Based pixel-prediction methods}
Developing deep Spiking Neural Networks (SNNs) for visual data processing can be achieved through two primary strategies \cite{li2023deep}. The first strategy involves transforming conventional Artificial Neural Networks (ANNs) into their SNN equivalents, whereas the second focuses on training SNNs from scratch using spike-based learning techniques. In the ANN-to-SNN conversion approach, pre-trained ANN models are adapted into spiking architectures. To regulate spike activity in classification tasks, Diehl et al. introduce a "layer-wise weight normalization" technique in combination with threshold tuning \cite{diehl2015fast}. Additionally, Yan et al. present a clamped and quantized training approach designed to reduce information loss during the transformation, ensuring minimal degradation in classification accuracy \cite{yan2021near}.

\begin{table*}[t!]
    \centering
    \renewcommand{\arraystretch}{1.3} 
    \caption{Comparison of between different SNN training strategies}
    \label{table:train_strategies}
    \setlength{\tabcolsep}{6pt} 
    \resizebox{\textwidth}{!}{
    \begin{tabular}{>{\centering\arraybackslash}p{3cm} | p{4cm} | p{4.5cm} | p{4.5cm} | p{4cm}}
    \hline
    
    \textbf{Training Strategies}  & \textbf{Key Methods} & \textbf{Advantages} & \textbf{Disadvantages} & \textbf{Use Cases} \\
    \hline \hline

    \multirow{2}{=}{\textbf{ANN to SNN Conversion}} & \textbullet\ Spike coding & \textbullet\ Utilizes pre-trained ANN models & \textbullet\ High latency  & \textbullet\ Deep learning tasks \\
     & \textbullet\ Layer-wise normalization & \textbullet\ Works well for deep networks & \textbullet\ Inefficient for real-time applications & ~ \\
    \hline

    \multirow{2}{=}{\textbf{Spike-based Backpropagation}} & \textbullet\ Surrogate Gradient Descent & \textbullet\ Enables direct SNN training & \textbullet\ Computationally expensive & \textbullet\ Supervised learning \\
    & \textbullet\ Backpropagation Through Time (BPTT) & \textbullet\ Achieves high accuracy & \textbullet\ Memory-intensive for long sequences & \textbullet\ Large-scale deep SNNs \\
    \hline

    \multirow{2}{=}{\textbf{Local Learning Rules}} & \textbullet\ Spike-Timing-Dependent Plasticity (STDP) & \textbullet\ Hardware-efficient & \textbullet\ Difficult to achieve high accuracy & \textbullet\ Unsupervised learning capabilities \\
    & \textbullet\ Hebbian Learning & \textbullet\ Suitable for neuromorphic hardware & \textbullet\ No direct supervision & \textbullet\ Low-power edge computing on neuromorphic chips \\
    \hline

    \multirow{3}{=}{\textbf{Evolutionary \& Reinforcement Learning}} & \textbullet\ Evolutionary Strategies (ES) & ~ & \textbullet\  Slow convergence & \textbullet\ Robotics \\
    & \textbullet\ Reinforcement Learning (RL) & \textbullet\ Suitable for decision-making & ~ & \textbullet\ Adaptive control \\
    & \textbullet\ Neuromodulated Plasticity & ~ & \textbullet\ High variance & \textbullet\ Decision-making tasks \\
    \hline

    \multirow{3}{=}{\textbf{Hybrid Learning}} & \textbullet\ STDP + Backpropagation & \textbullet\ Balances computational efficiency and accuracy & \textbullet\ Increased complexity & \textbullet\ Optimization across multiple tasks \\
    & \textbullet\ ANN-SNN Hybrid Training & \textbullet\ Adaptable to different applications & \textbullet\ Requires fine-tuning & ~ \\
    & \textbullet\ Meta-Learning for SNNs & ~ & ~ & ~ \\
    \hline
    \end{tabular}}
\end{table*}

The direct training of SNN can be accomplished through supervised or unsupervised learning paradigms. Within the scope of unsupervised learning, the STDP strategy is recognized as a foundational approach. The method has been applied in various tasks, such as training networks for image classification \cite{diehl2015unsupervised} and object recognition using event-driven data representations \cite{liu2020unsupervised}. Additionally, STDP has been integrated with dynamic threshold adjustments to enhance recognition capabilities in video-based facial disguise detection \cite{liu2019deep}. Supervised learning in SNNs presents unique challenges due to the discrete nature of spike-based neuron activations, making direct gradient computation difficult. To address this, \cite{lee2016training} introduces an approach that smooths spike events, enabling gradient propagation across network layers using chain rule-based computations. A more structured approach, "Spatial-Temporal Backpropagation (STBP)," is leveraged for image classification tasks by incorporating surrogate gradients to approximate non-differentiable spike-based updates \cite{wu2018spatio}. A recent advancement explores a hybrid learning framework that merges STDP with STBP to enhance object detection capabilities in SNNs \cite{chakraborty2021fully}. This integration aims to leverage the efficiency of unsupervised feature extraction while benefiting from the structured optimization of supervised learning techniques.The summary of different SNN training strategies is detailed in Table \ref{table:train_strategies}.

Roy et al. \cite{roy2019synthesizing} introduced a method to synthesize images from multiple modalities within a spike-based representation. They utilized a spiking autoencoder to encode inputs into compact spatio-temporal representations, which were then decoded for image reconstruction. The training process involved computing the membrane potential loss at the output layer and backpropagating it using a sigmoid approximation of the neuron's activation function to ensure differentiability. Later, in 2021, a spiking autoencoder with temporal coding and pulse-based training was presented \cite{comcsa2021spiking}, demonstrating the importance of inhibition for memorizing inputs over extended periods before generating the expected output spikes. 

Castagnetti et al. \cite{castagnetti2023spiden} introduced the first SNN-driven Gaussian image denoising model, employing approximate gradient optimization and temporal backpropagation (BPTT) techniques to train the network within the spiking framework. In 2024, two advancements were introduced. A spiking U-Net using multi-threshold neurons was proposed, leveraging an ANN-SNN adaptation framework followed by refinement using previously trained U-Net networks \cite{li2024deep}. Additionally, a non-training-based method was proposed, utilizing threshold guidance for SNN-based diffusion models in Gaussian image denoising \cite{cao2024spiking}. Additionally, ESDNet \cite{song2024learning}, an image-deraining model, introduced a spiking residual block that converted inputs into spike signals, optimizing membrane potentials adaptively with attention weights to mitigate information loss from discrete binary activations. While these methods have shown promise in image reconstruction and denoising, they are limited to low-resolution images and Gaussian noise removal. This makes them inadequate to use for high-resolution, complex underwater image sequences with non-linear haze distributions.

\subsubsection{Conversion of continuous pixel intensities to spikes}
Converting continuous-valued input signals to a sparse, spike-based representation without significant information loss is one of the most important tasks in any SNN-based computing paradigm. Several spike-coding-decoding techniques have been proposed to tackle the current lack of neuromorphic sensing for robotics \cite{dupeyroux2022toolbox}. Temporal coding \cite{kiselev2016rate} uses spike timing that is inversely proportional to pixel intensity. The neuron that receives the highest activation value fires a spike first compared to the neuron that receives a lower value. Since there is only one spike firing at each timestep, the average spike rate remains lower, requiring larger timesteps to capture relevant information \cite{hwang2024one}. 

In phase coding, temporal information is encoded into spike representations using a global oscillator \cite{kim2018deep}. In binary phase coding, a different weight determined by the power of \(2\) is assigned to each timestep. Thus, the activation values are encoded by both the spike pattern and spike rates.  These methods are not suited for architectures where network size and dataset size are large \cite{kim2022rate}. Rate coding \cite{auge2021survey} is the widely used spike-coding technique in classification tasks, where the input value is encoded into a spike train over a pre-defined number of timesteps. The number of generated spikes, sampled from the Poisson distribution, is proportional to the input magnitude. In the direct coding technique \cite{mukhoty2023direct}, the continuous-valued input is fed directly to the first learning-based layer of the network, which outputs the weighted float-point values. The subsequent spiking neuron layer repeatedly receives these values over the pre-defined timesteps to generate spikes. 

\section{Knowledge Background}
\subsection{Spiking Neuron with Adaptive Threshold}
The leaky integrate-and-fire (LIF) neuron dynamics with a modified adaptive threshold mechanism are used in the spike coding module along with the convolutional layers. The neuron dynamics of an LIF neuron are modeled using an RC circuit \cite{eshraghian2023training} and is represented as: 
\begin{equation}
    \centering
    \tau \frac{d V(t)}{dt}=-V(t)+I(t)R.
    \label{eq:c2_lif_dynamics}
\end{equation}
Here, \(\tau\) represents the time constant (\(\tau=RC\)), with \(R\) and \(C\) denoting the resistive and capacitive components of the RC circuit, respectively. The function \(I(t)\) corresponds to the applied input current. For a constant input \(I(t)=I\), the solution to (\ref{eq:c2_lif_dynamics}) is:

\begin{figure}[h!]
    \centering
    \includegraphics[width=0.8\columnwidth]{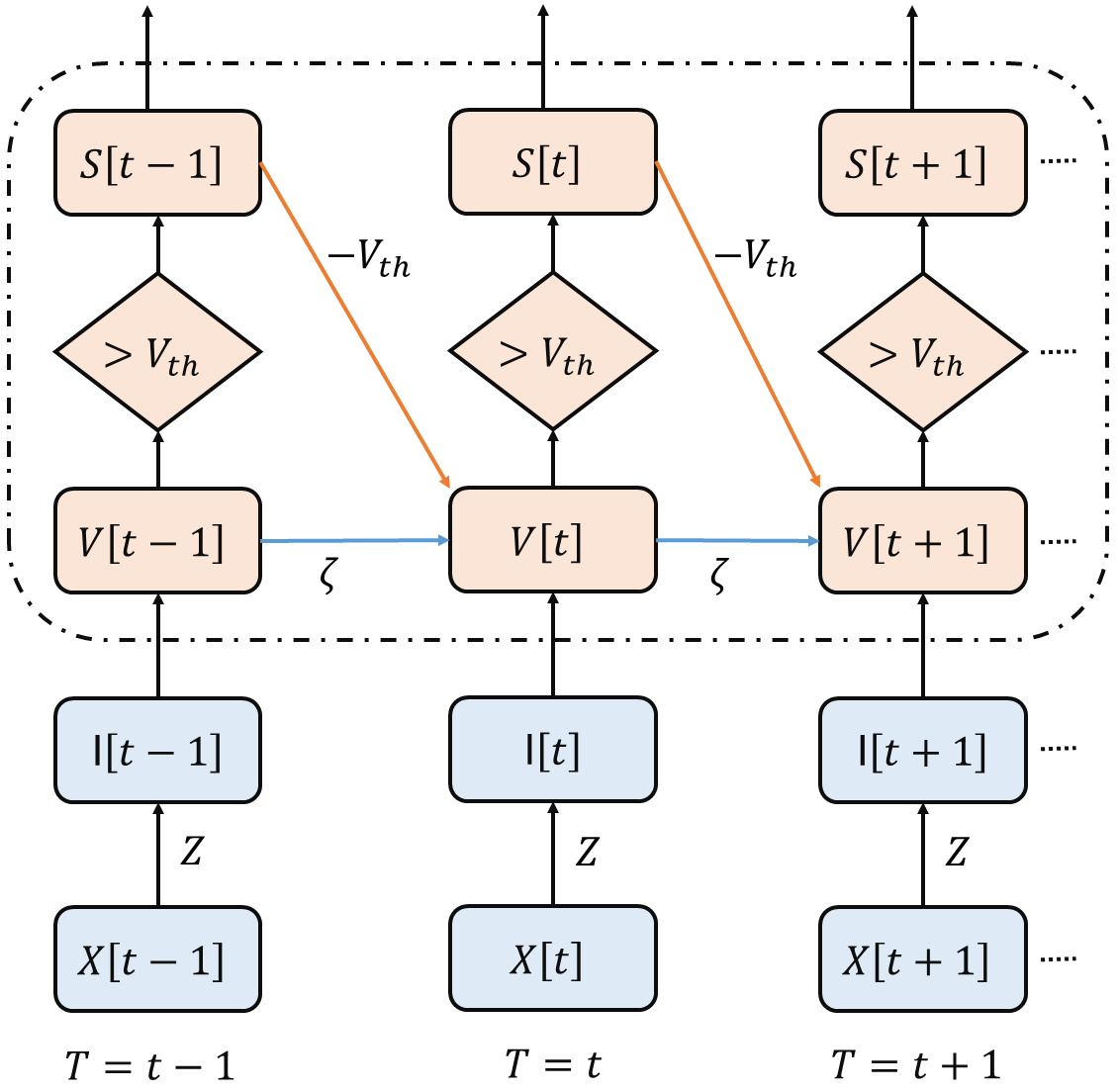}
    \caption{Computational graph of LIF neuron over the timesteps.}
    \label{fig:c2_lif_computation}
\end{figure}

\begin{equation}
    \centering
    V(t)=IR+\left[ V_{0}-IR \right]e^{-t/\tau}.
    \label{eq:c2_analytical_soln}
\end{equation}

To make this model compatible with sequence-based neural networks, we discretize time \(t\to t+1 \) with a small step \(\Delta t\). Applying the Euler method to (\ref{eq:c2_lif_dynamics}):

\begin{equation}
    \centering
    \frac{V(t+\Delta t)-V(t)}{\Delta t}= -\frac{V(t)}{\tau} + \frac{IR}{\tau}.
    \label{eq:c2_euler1}
\end{equation}
 
Rearranging (\ref{eq:c2_euler1}) for \(V(t+\Delta t)\):
\begin{equation}
    \centering
    V(t+\Delta t)=V(t)+\frac{\Delta t}{\tau}\left( -V(t) + IR\right).
    \label{eq:c2_euler2}
\end{equation}

Assuming \(\Delta t = 1\), we define the decay rate \(\zeta = e^{-1/\tau}\). Thus, the update equation simplifies to:
\begin{equation}
    \centering
    V[t]=\zeta V[t-1]+(1-\zeta)I[t].
    \label{eq:c2_approximate_soln}
\end{equation}

The above formula is the Euler-approximated discretized form of the LIF dynamic equation. In the learning-based frameworks, the weighting factor \((1- \zeta)\) of the input \(I[t]\) is considered as a learnable parameter. Instead of directly using \(I[t]\), we define \(I[t] = Z X[t]\). \(X[t]\)  represents the neuron's received signal, while \(Z\) denotes the adaptive synaptic weight acquired through learning. Therefore, 

\begin{equation}
    \centering
    V[t]=\zeta V[t-1]+Z X[t].
\end{equation}

Neurons exhibit a spiking behavior. That is, if the internal potential \(V\) exceeds a defined threshold \(V_{th}\), it triggers a spike firing, followed by a reset of the neuron's state. The spiking event follows the formulation \(S[t]=\Theta (V[t]-V_{th})\). Here \(\Theta (V[t]-V_{th})\) is the Heaviside function, which is defined as:

\begin{equation}
    \Theta(V[t]-V_{th}) = 
    \begin{cases}
    1, & V[t] \ge V_{th}. \\
    0, & otherwise.
    \end{cases}
\end{equation}

To enforce the reset after a spike, we subtract the threshold potential weighted by the spike output:

\begin{equation}
    \centering
    V[t]=\zeta V[t-1]+Z X[t]-S[t-1]V_{th}.
\end{equation}

In the direct coding scheme with an adaptive threshold, the threshold membrane potential \(V_{th}\), is considered as learnable parameter. 

\subsection{Training Strategies}\label{sec:BPTT}
Unlike the widely adopted gradient descent optimization used for training conventional ANNs \cite{javanshir2022advancements}, optimizing SNNs presents significant challenges because spike-based computations are inherently non-differentiable. The three main approaches are: (i) conversion from ANN to SNN, where a conversion from formal to event-driven input can lead to a large loss of accuracy and poor interpretability; ii) supervised learning, in which various approaches aim to address the fundamental limitation of spike-based mathematical expressions-specifically, their non-differentiability, which is essential for any backpropagation algorithm. iii) unsupervised methods, such as "Spike-Timing-Dependent Plasticity (STDP)", in which synaptic weight adjustments are influenced by the temporal relationship of spike events from pre- and post-synaptic neurons. (If the timing is short, this implies causality and large weight and vice versa). The last training procedure, very close to its biological counterpart, is \textit{local training}. The local training strategy is characterized by high \textit{adaptability} to the changes of inputs and environments, quite differently from traditional neural networks.

In reference to Fig.\ref{fig:c2_lif_computation}, the backpropagation algorithm in SNN-based model at a single timestep is formulated.

\begin{equation}
    \frac{\partial \mathcal{L}}{\partial Z}=\frac{\partial \mathcal{L}}{\partial S}\frac{\partial S}{\partial V}\frac{\partial V}{\partial I}\frac{\partial I}{\partial Z}.
    \label{eq:BP_SNN}
\end{equation}
The output spike generated during the forward propagation is given as:
\begin{equation}
    S[t]= \Theta (V[t]-V_{th}) = 
    \begin{cases}
    1, & V \ge V_{th}. \\
    0, & otherwise.
    \end{cases}
    \label{eq:forward}
\end{equation}
The derivative calculated during the backward propagation for (\ref{eq:forward}) is:
\begin{equation}
    \frac{\partial S}{\partial V[t]}= \delta (V-V_{th}) = 
    \begin{cases}
    +\infty , & V[t] = V_{th}. \\
    0, & V[t] \neq  V_{th}.
    \end{cases}
\end{equation}
where, \(\delta(.)\) is the Dirac-Delta function. 

This implies that the term \(\frac{\partial S}{\partial V}\) will almost always be zero, thereby inhibiting the learning process. In this case, the network training will be unstable with the use of \(\delta(.)\) function to calculate the gradient and apply the gradient descent. 

The surrogate gradient-based method with fast-sigmoid function is used in this work to mitigate the dead neuron problem \cite{eshraghian2023training}. The threshold-centered membrane potential is defined as \(\tilde{V} = V[t]-V_{th}\). The fast-sigmoid function and its derivative used for backpropagation is represented as:

\begin{equation}
    \tilde{S} \approx \frac{\tilde{V}}{(1+\lambda\left| \tilde{V} \right|)}.
\end{equation}

\begin{equation}
    \frac{\partial \tilde{S}}{\partial V} \approx \frac{1}{(1+\lambda\left| \tilde{V} \right|)^{2}}.
\end{equation}
where \(\lambda\) modulates the smoothness of the surrogate function.

Since SNNs operate in the temporal domain, the weight \(Z\) influences the membrane potential and loss across all time steps. Since the weights are shared across all timesteps (\(Z\) remains unchanged for \(Z[0], Z[1],...,Z[T]\)), the global loss gradient \(\mathcal{L}\) computed over the weight \(Z\) for all past timesteps \( p\le t\) is defined as:

\begin{equation}
    \frac{\partial \mathcal{L}}{\partial Z}=\sum_{t=1}^{T}\frac{\partial \mathcal{L}[t]}{\partial Z}=\sum_{t=1}^{T}\sum_{ p\le t}^{}\frac{\partial \mathcal{L}[t]}{\partial Z[p]}\frac{\partial  Z[p]}{\partial Z}= \sum_{t=1}^{T}\sum_{ p\le t}^{}\frac{\partial \mathcal{L}[t]}{\partial Z[p]}.
\end{equation}

For instance, if only the immediate prior influence corresponding to timestep \(p=t-1\) is taken into account, the backward pass needs to be traced back in time by a single step. In this case, the influence of \(Z[t-1]\) on \(\mathcal{L}[t]\) is described as:

\begin{equation}
    \frac{\partial \mathcal{L}[t]}{\partial Z[t-1]}=\frac{\partial \mathcal{L}[t]}{\partial S[t]}\frac{\partial \tilde{S}[t]}{\partial V[t]}\frac{\partial V[t]}{\partial V[t-1]}\frac{\partial V[t-1]}{\partial I[t-1]}\frac{\partial I[t-1]}{\partial Z[t-1]}.
\end{equation}

Here, \(\frac{\partial V[t]}{\partial V[t-1]}\) denotes the decay value defined for a LIF neuron; \(\frac{\partial V[t-1]}{\partial I[t-1]} = 1\), and \(\frac{\partial I[t-1]}{\partial Z[t-1]}\) is the presynaptic input.

By combining surrogate gradients with BPTT, the training of SNNs becomes both feasible and efficient. The surrogate gradient approximations enable non-zero gradients even if the neuron's internal voltage approaches its activation limit. The BPTT strategy ensures that the temporal dependencies are captured across timesteps. The surrogate gradient used in this work aligns with established methods in the literature \cite{eshraghian2023training}, and provides smooth gradient flow while preserving the temporal dynamics essential for SNNs.

\subsection{Spike coding of continuous-values Signals}
\begin{figure}[h!]
\includegraphics[width=\columnwidth]{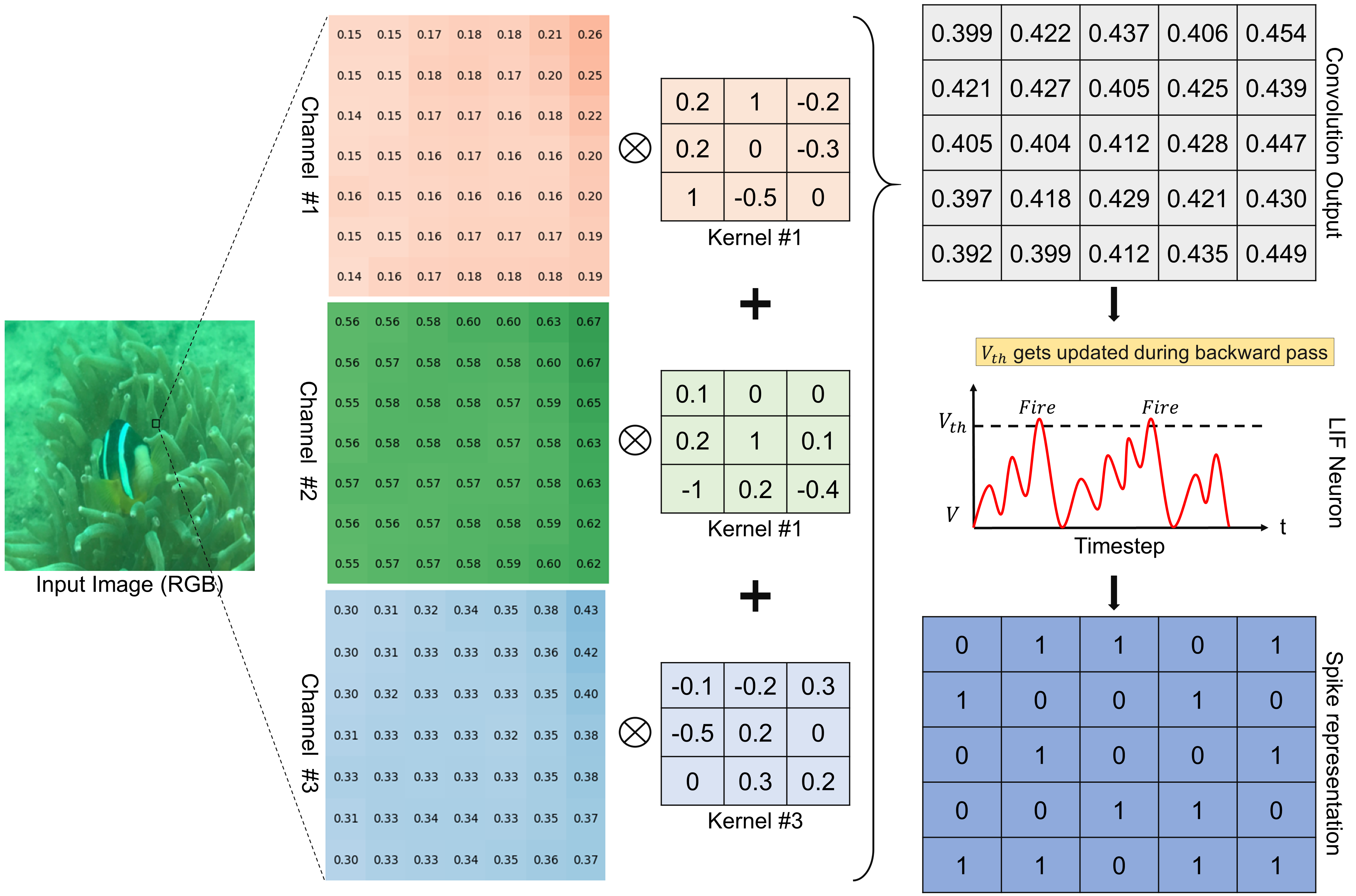}
\centering
\caption{Spike coding using LIF neuron with adaptive threshold: In the first stage, feature maps are generated when each channel of the input RGB images, and undergoes convolution with respective kernel. The output is then fed to a spiking neuron, where the membrane potential (\(V\)) is updated depending on the presynaptic input. When \(V\) exceeds the threshold membrane potential (\(V_{th}\)), a spike is fired, and \(V\) resets to a predefined value. In the adaptive threshold mechanism, \(V_{th}\) is updated during backpropagation.}
\label{fig:lif_adaptive}
\end{figure}

A LIF neuron with an adaptive threshold mechanism is used after the convolutional input layer for the spike coding of continuous input values. For processing the visual inputs, a 2D convolutional layer followed by the adaptive LIF neuron converts the continuous pixel intensity values to their equivalent binary spike representation. In the decoding phase, these spikes are then fed to the output convolutional layer followed by a LIF neuron whose membrane potential, without resetting, is used directly to compute the reconstruction loss. The entire spike-coding process of a single image during a single timestep is presented in Fig.\ref{fig:lif_adaptive}.

\subsubsection{Energy Measurement}
The energy consumption for CNNs and SNNs are computed based on the total number of Floating Point Operations (FLOPs) and  Synaptic Operators (SOPs) respectively. The energy consumption is based on the standard CMOS technology \cite{horowitz20141} as shown in Table  \ref{table:energy_cmos}. 

\begin{table}[H]
\renewcommand{\arraystretch}{1.3} 
\caption{Energy consumption for 45nm CMOS process.}
\label{table:energy_cmos}
\centering
\begin{tabular}{l|c}
\hline
\bfseries Floating Point (FP) Operation & \bfseries Energy (pJ) \\ 
\hline \hline
32 bit FP MULT                         & 3.7                  \\ 
32 bit FP ADD                          & 0.9                  \\ 
32 bit FP MAC (\(E_{MAC}\))            & 4.6                  \\ 
32 bit FP ACC (\(E_{ACC}\))              & 0.9                  \\ 
\hline
\end{tabular}
\end{table}

The FLOPs and SOPs for each convolution layer \(l\) in CNN and SNN are computed as follows:

\begin{equation}
    \centering
    \begin{matrix}
    CNN: & FLOPs_{CNN}(l)=l_{Cin} \times l_{Cout} \times l_{k}^{2} \times l_{l} \times l_{w}\\
    & \\
    SNN: & SOPs_{SNN}(l)=FLOPs_{CNN}(l) \times S_{r}(l)
\end{matrix}
\end{equation}
where, \(l_{Cin}\) is the number of input channels at layer \(l\), \(l_{Cout}\) is the number of output channels at layer \(l\), \(l_{k}\) is the size of the kernel at layer \(l\), \(l_{l}\) is the length of the output feature map at layer \(l\), \(l_{w}\) is the width of the output feature map at layer \(l\) and \(S_{r}(l)\) is the spike rate \cite{kim2022beyond} at each SNN layer \(l\).

\begin{equation}
    \centering
    \begin{matrix}
    S_{r}(l) & =\frac{Number\;of\;spikes\;over\;all\;timesteps}{Number\;of\;neuron}     \\
     & \\
     & = \frac{\sum_{c=1}^{l_{Cout}}\sum_{i=1}^{l_{l}}\sum_{j=1}^{l_{w}}\sum_{t=1}^{T}S^{t}_{l}(i,j,c)}{l_{Cout}\times l_{l} \times l_{w}}
    \end{matrix}
\end{equation}

The spike rate \(S_{r}(l)\) effectively scales the computational cost of SNNs by accounting for spike-driven sparse computations. Since SNNs operate on binary spikes rather than continuous activations, the overall number of operations can be significantly reduced compared to traditional CNNs. 

Given the operation counts per layer, the total inference energy for CNNs and SNNs across all layers can be determined as follows:
\begin{equation}
    \centering
    \begin{matrix}
CNN: & E_{CNN}=\sum_{l=1}^{L}FLOPs_{CNN}(l) \times E_{MAC}\\ 
 & \\ 
SNN: & E_{SNN}=\sum_{l=1}^{L}SOPs_{SNN}(l) \times E_{ACC}
\end{matrix}
\end{equation}
where \(L\) is the total number of layers in the architecture, \(E_{MAC}\) represents the energy per MAC operation, and \(E_{ACC}\) represents the energy per accumulation operation. \(E_{MAC}\) and  \(E_{ACC}\) can be obtained from Table  \ref{table:energy_cmos}.

The percentage of energy reduction is calculated as:

\begin{equation}
    \centering
    \Delta E (\%) =\frac{E_{CNN}-E_{SNN}}{E_{CNN}} \times 100
\end{equation}

\section{Modelling of snnTrans-DHZ Architecture}

\begin{figure*}[t!]
\includegraphics[width=\textwidth]{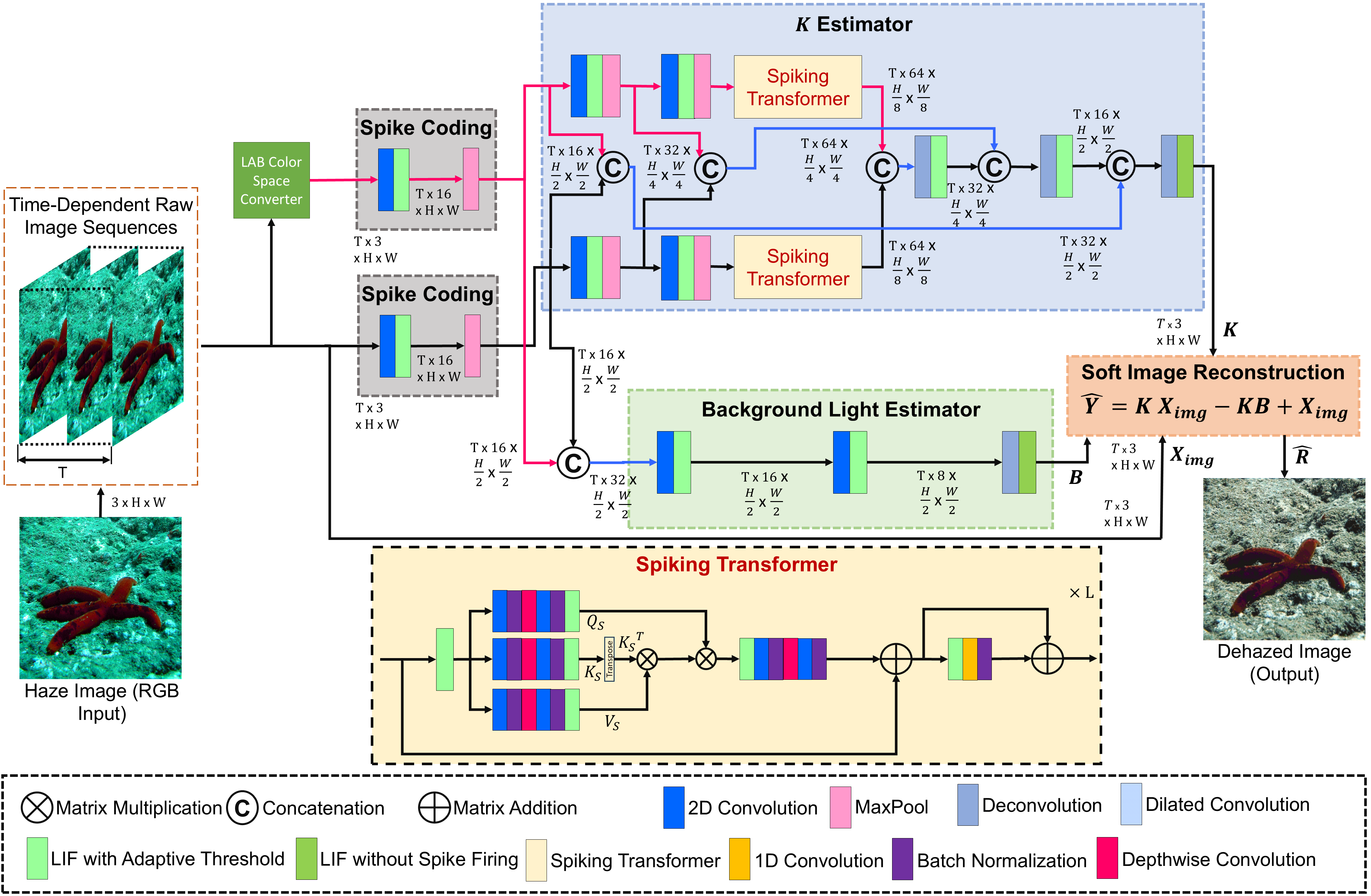}
\centering
\caption{Schematic representation of snnTrans-DHZ framework for underwater image dehazing. The spike coding is performed simultaneously on the RGB and LAB color space converted time-dependent raw image sequences. In the \textit{K estimator} module, the low-level features from RGB and LAB color spaces are extracted separately using the spike transformer. They are concatenated and fed to the deconvolution layer. The separately extracted feature maps at different spatial resolutions are fused and fed to the corresponding deconvolution layers. In the \textit{background light estimator} module, RGB-LAB spike-coded signals are concatenated and fed to the convolution layers, followed by ALIF neurons for feature extraction. The \(\textbf{K}\) and \(\textbf{B}\) estimates, together with \(\textbf{X}^{RGB}_{img}\), are fed to the \textit{soft image reconstruction} module for the reconstruction of haze-free images. The ALIF neurons with a learnable threshold membrane potential are used throughout the network.}
\label{fig:c4_snntrans_arc}
\end{figure*}

The schematic representation of the proposed snnTrans-DHZ framework for underwater image dehazing is presented in Fig.\ref{fig:c4_snntrans_arc}. The raw underwater images are first converted into time-dependent image sequences by repeatedly passing the static image to a user-defined timestep value \(T\). The RGB sequences are then converted into LAB color space representations and processed simultaneously. The \textit{K estimator} utilizes spiking transformer-based architecture along with convolutional SNN layers to extract features from different color space representations. A joint feature upsampling is performed using the deconvolution layers, which leads to the extraction of \(\textbf{K}\). The time-dependent image sequences in RGB and LAB color spaces are concatenated and fed to the convolutional SNN layers to jointly estimate the background light estimate \(\textbf{B}\). Finally, the \textit{soft image reconstruction} module reconstructs the haze-free, visibility-enhanced image \(\hat{\textbf{R}}\) by utilizing \(\textbf{K}\), \(\textbf{B}\) and \(\textbf{X}^{RGB}_{img}\).  

\subsection{Adaptive Leaky Integrate-and-Fire (ALIF) Neuron}
The Adaptive Leaky Integrate-and-Fire (ALIF) neuron modifies the standard LIF model by replacing the fixed threshold with a learnable threshold \(V_{th}\). It allows the ALIF neuron to dynamically adjust its firing behavior based on past activity. Unlike the LIF neuron with a fixed threshold membrane potential, in ALIF neurons, the \(V_{th}\) is treated as a learnable parameter optimized through gradient descent. This adaptation eliminates the need for tedious selection of an optimal \(V_{th}\). By making \(V_{th}\) adaptive, the ALIF neuron enhances temporal coding and learning capabilities, making it more biologically plausible and efficient for SNN-based applications.

\subsection{RGB to LAB Transformation}
The LAB representation (CIELAB) offers a perceptually consistent color representation, closely aligning with human visual perception to ensure accurate color differentiation. It consists of three components: L (lightness), A (green-red), and B (blue-yellow), providing a device-independent representation of color. This conversion enhances robustness in various computer vision tasks by reducing sensitivity to illumination changes. The conversion from RGB to LAB is performed through an intermediate XYZ color space. Given an RGB image sequence \(\tilde{\textbf{X}}^{RGB}_{img}\), each pixel's linear RGB values \((R,G,B)\) are first normalized to \([0,1]\) and then converted to CIEXYZ using the standard sRGB transformation matrix \cite{rao2022developing}.
\begin{equation} 
    \left[ \begin{matrix}
    X \\
    Y \\
    Z \\
    \end{matrix} \right]= 
    \left[ \begin{matrix}
    0.412453 & 0.357580 & 0.180423 \\
    0.212671 & 0.715160 & 0.072169 \\
    0.019334 & 0.119193 & 0.950227 \\
    \end{matrix} \right]\left[ \begin{matrix}
    R \\
    G \\
    B \\
    \end{matrix}  \right]
\end{equation}

For gamma correction, \(\left\{ R,G,B \right\}\) values are first linearized \cite{leon2006color} using:
\begin{equation}
    \centering
    C_{linear} = \begin{cases} \frac{C}{12.92} & ,C \leq 0.04045 \\
    \left( \frac{C+0.055}{1.055} \right)^{2.4} & ,C > 0.04045 \end{cases} 
\end{equation}
where \(C \in \left\{ R,G,B \right\}\). The \(\left\{ X,Y,Z \right\}\) values are mapped to LAB color space using the following nonlinear transformation:

\begin{equation}
    \centering
    f(u) = \begin{cases} u^{\frac{1}{3}} & ,u > 0.008856 \\
 7.787u + \frac{4}{29} & ,u \leq 0.008856 \end{cases}
\end{equation}
where, \(u\) is \(\frac{X}{X_{n}}\), \(\frac{Y}{Y_{n}}\), or \(\frac{Z}{Z_{n}}\). The LAB components are then computed as:
\begin{equation}
    \centering
    \begin{matrix}
    L= &  116  f\left( \frac{Y}{Y_{n}} \right) - 16 \\
    A= &  500\left( f\left( \frac{X}{X_{n}} \right)-f\left( \frac{Y}{Y_{n}} \right) \right)\\
    B= &  200\left( f\left( \frac{Y}{Y_{n}} \right)-f\left( \frac{Z}{Z_{n}} \right) \right)
    \end{matrix}
\end{equation}
where, \(\left(  X_{n},Y_{n},Z_{n}  \right)\) represent the white achromatic reference illuminant. Since the input image is time-dependent, we apply the conversion frame-wise to the image sequence: 
\begin{equation}
    \centering
    \tilde{\textbf{X}}^{LAB}_{img}=RGB\to LAB\left( \tilde{\textbf{X}}^{RGB}_{img} \right)
\end{equation}
where \(\tilde{\textbf{X}}^{LAB}_{img}\) represents the LAB image sequences over \(T\) timesteps.

\subsection{Model Description}
 The raw RGB input image \(\textbf{X}^{RGB}_{img} \in \mathcal{R}^{3\times H \times W}\), is first converted to time-dependent image sequences \(\tilde{\textbf{X}}^{RGB}_{img} \in \mathcal{R}^{T\times 3\times H \times W}\). RGB image sequences \(\tilde{\textbf{X}}^{RGB}_{img}\) are then converted to \(\tilde{\textbf{X}}^{LAB}_{img} \in \mathcal{R}^{T\times 3\times H \times W}\). The continuous valued time-dependent input image sequences \(\tilde{\textbf{X}}^{RGB}_{img}\) and \(\tilde{\textbf{X}}^{LAB}_{img}\) are processed through two distinct spike coding modules, transforming them into spike-based representations. The spike-coded RGB and LAB branch outputs after downsampling are given by \(\textbf{S}_{RGB} \in \left\{ 0,1 \right\}^{T\times 16\times \frac{H}{2} \times \frac{W}{2}}\) and \(\textbf{S}_{LAB} \in \left\{ 0,1 \right\}^{T\times 16\times \frac{H}{2} \times \frac{W}{2}}\).
 
\subsubsection{Background Light Estimator}
This module estimates the background light \(\textbf{B}\) from the concatenated spike‐coded features of the RGB and LAB pathways. For the concatenated RGB and LAB spike-coded outputs along the channel dimension, be \(\textbf{X}_{BL} = concat(\textbf{S}_{RGB}, \textbf{S}_{LAB}) \in \left\{ 0,1 \right\}^{T\times 32\times \frac{H}{2} \times \frac{W}{2}}\). Upon sequential convolution and deconvolution layers followed by the ALIF neuron, the \(\textbf{B}\) is obtained by:

\begin{equation}
\begin{split}
    \textbf{B}= & LIF_{final}(DeConv(ALIF(Conv(ALIF( \\
                & Conv(\textbf{X}_{BL})))))) \in \mathcal{R}^{T \times 3 \times H \times W}.
\end{split}
\end{equation}

\subsubsection{K Estimator}
The goal of this module is to estimate the spatially varying transmission function \(\textbf{K} \in \mathcal{R}^{T \times 3 \times H \times W}\). It comprises dual encoder streams, with one handling RGB spike-coded features while the other processes LAB spike-coded features and a single decoder that upsamples the combined features to produce the final \(\textbf{K}\) estimate.

There are two encoding stages for the RGB and LAB branches, followed by a spiking transformer module. Each encoding stage consists of a convolution layer, followed by an ALIF neuron layer and a downsampling block to reduce the spatial dimension.

Encoder stage 1: 
\begin{equation}
\begin{split}
    \textbf{S}^{K}_{RGB,e1} = & MaxPool(ALIF( \\
    & Conv(\textbf{S}_{RGB}))) \in \left\{ 0,1 \right\}^{T \times 32 \times \frac{H}{4} \times \frac{W}{4}}.
\end{split}
\end{equation}

\begin{equation}
\begin{split}
    \textbf{S}^{K}_{LAB,e1} = & MaxPool(ALIF( \\
    & Conv(\textbf{S}_{LAB}))) \in \left\{ 0,1 \right\}^{T \times 32 \times \frac{H}{4} \times \frac{W}{4}}.
\end{split}
\end{equation}

Encoder stage 2: 

\begin{equation}
\begin{split}
    \textbf{S}^{K}_{RGB,e2} = & MaxPool(ALIF( \\
    & Conv(\textbf{S}^{K}_{RGB,e1}))) \in \left\{ 0,1 \right\}^{T \times 64 \times \frac{H}{8} \times \frac{W}{8}}.
\end{split}
\end{equation}

\begin{equation}
\begin{split}
    \textbf{S}^{K}_{LAB,e2} = & MaxPool(ALIF( \\
    & Conv(\textbf{S}^{K}_{LAB,e1}))) \in \left\{ 0,1 \right\}^{T \times 64 \times \frac{H}{8} \times \frac{W}{8}}.
\end{split}
\end{equation}

\begin{figure*}[h!]
    \centering
    \includegraphics[width=0.9\textwidth]{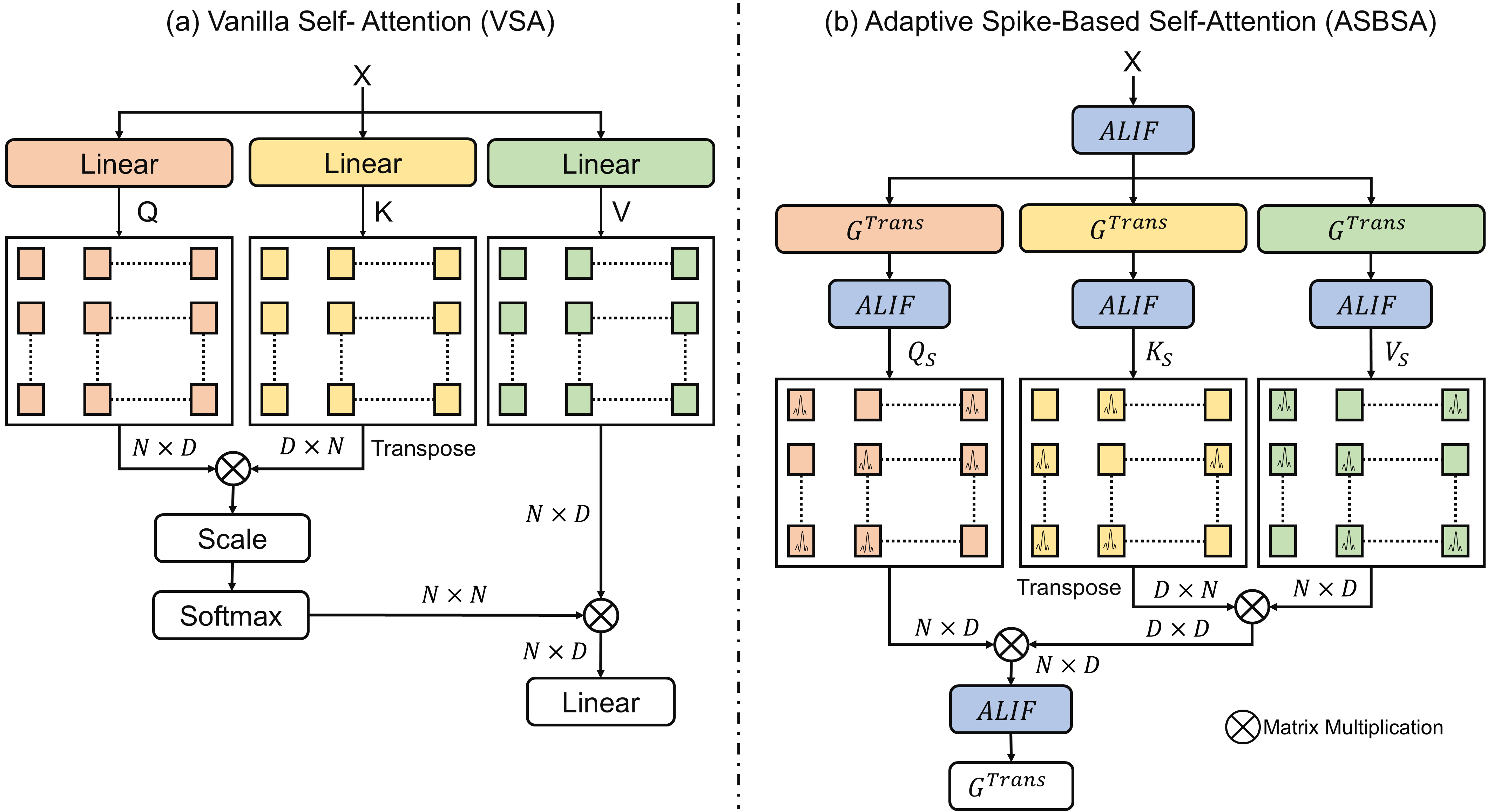}
    \caption{Key differences of "Vanilla Self-Attention (VSA)" and Adaptive Spike-Based Self-Attention (ASBSA). (a) VSA follows the traditional attention mechanism, where input \(X\) is mapped into "query (\(Q\)), key (\(K\)), and value (\(V\))" representations, followed by scaled dot-product attention and softmax normalization; (b) SBSA introduces ALIF neurons with a learnable threshold potential and spike-based transformation module \(G^{Trans}\) replacing traditional linear layers. This enhances energy efficiency and sparsity by leveraging spiking neural dynamics. The attention computation retains matrix multiplications but eliminates the scale and softmax operations.}
    \label{fig:c4_vsa_asbsa}
\end{figure*}

\paragraph{Spiking Transformer Blocks}

A series of transformer blocks is applied to the downsampled features for each branch. One transformer block \(\mathcal{B}\) comprises two main parts: a spiking self-attention module and an MLP block with residual connections. Unlike the "Vanilla Self-Attention (VSA)" \cite{vaswani2017attention}, a group of convolution and batch normalization operations is performed to extract spike-based query \((\textbf{Q}_{S})\), key \((\textbf{K}_{S})\), and value \((\textbf{V}_{S})\) matrix from the input of the transformer block. The key differences of VSA and Adaptive Spike-Based Self-Attention (ASBSA) is presented in Fig. \ref{fig:c4_vsa_asbsa}.  The group operations in the transformer module, \(G^{Trans}\), consist of a convolution layer, followed by batch normalization, depth-wise Convolution (DWConv) layer, convolution layer, and batch normalization. The convolution layer defined here is two-dimensional. Otherwise, it will be explicitly mentioned.

For the extraction of spike-based query \((\textbf{Q}_{S})\), key \((\textbf{K}_{S})\), and value \((\textbf{V}_{S})\) for a given input \(\textbf{X}^{Trans}_{in} \in \mathcal{R}^{T \times 64 \times \frac{H}{8} \times \frac{W}{8}}\):

\begin{equation}
\begin{split}
    \textbf{Q}_{S}= & ALIF(G^{Trans}( \\
    & ALIF(\textbf{X}^{Trans}_{in})) \in \left\{ 0,1 \right\}^{T \times N \times D}.
\end{split}
\end{equation}

\begin{equation}
\begin{split}
    \textbf{K}_{S}= & ALIF(G^{Trans}( \\
    & ALIF(\textbf{X}^{Trans}_{in})) \in \left\{ 0,1 \right\}^{T \times N \times D}.
\end{split}
\end{equation}

\begin{equation}
\begin{split}
    \textbf{V}_{S}= & ALIF(G^{Trans}( \\
    & ALIF(\textbf{X}^{Trans}_{in})) \in \left\{ 0,1 \right\}^{T \times N \times D}.
\end{split}
\end{equation}

where \(N\) is the token number, and \(D\) represents the dimensionality of embedded vector. 

The interaction between \(\textbf{Q}_{S},\textbf{K}_{S},\textbf{V}_{S}\) within the ASBSA is formulated as:

\begin{equation}
    \begin{split}
        ASBSA(\textbf{Q}_{S},\textbf{K}_{S},\textbf{V}_{S})= & \\
        & ALIF(\textbf{Q}_{S}(\textbf{K}^{T}_{S}\textbf{V}_{S})) \\
        & \in \left\{ 0,1 \right\}^{T \times N \times D}.
    \end{split}
\end{equation}

The input \(\textbf{X}^{Trans}_{mlp}\) to the transformer linear layer is:

\begin{equation}
    \begin{split}
        \textbf{X}^{Trans}_{mlp} = & G^{Trans}(ASBSA(\textbf{Q}_{S},\textbf{K}_{S},V_{S})) + \\
        & \textbf{X}^{Trans}_{in}  \in \mathcal{R}^{T \times 64 \times \frac{H}{8} \times \frac{W}{8}}.
    \end{split}
\end{equation}

The output of the spike-based transformer block is:
\begin{equation}
\begin{split}
    \textbf{X}^{Trans}_{out} = & \textbf{X}^{Trans}_{mlp} + BN(Conv1d( \\
    & ALIF(\textbf{X}^{Trans}_{mlp})) \in \mathcal{R}^{T \times 64 \times \frac{H}{8} \times \frac{W}{8}}.
\end{split}
\end{equation}

Considering that the RGB and LAB pathway inputs to the spiking transformer modules are \(\textbf{S}^{K}_{RGB,e2}\) and \(\textbf{S}^{K}_{LAB,e2}\) generate the output as \(\textbf{S}^{K}_{RGB,trans}\) and, \(\textbf{S}^{K}_{LAB,trans}\) respectively. The feature maps at the corresponding spatial resolution are concatenated and fed to the decoder layers, which finally outputs \(\textbf{K} \in \mathcal{R}^{T \times 3 \times H \times W}.\)

Output at decoder stage 1:
\begin{equation}
    \begin{split}
        \textbf{S}^{K}_{d1} = & ALIF(DeConv(concat(\textbf{S}^{K}_{RGB,trans}, \\
        & \textbf{S}^{K}_{LAB,trans}))) \in \left\{ 0,1 \right\}^{T \times 32 \times \frac{H}{4} \times \frac{W}{4}}.
    \end{split}
\end{equation}

Output at decoder stage 2:
\begin{equation}
    \begin{split}
        \textbf{S}^{K}_{d2} = & ALIF(DeConv(concat(\textbf{S}^{K}_{d1},concat(\textbf{S}^{K}_{RGB,e1}, \\ & \textbf{S}^{K}_{LAB,e1})))) \in \left\{ 0,1 \right\}^{T \times 16 \times \frac{H}{2} \times \frac{W}{2}}.
    \end{split}
\end{equation}

Output at decoder stage 3:
\begin{equation}
    \begin{split}
        \textbf{K} = & LIF_{final}(DeConv(concat(\textbf{S}^{K}_{d2}, \\
        & concat(\textbf{S}_{RGB}, \textbf{S}_{LAB})))) \in \mathcal{R}^{T \times 3 \times H \times W}.
    \end{split}
\end{equation}

\subsubsection{Soft Image Reconstruction}
The underwater image formation \cite{jaffe1990computer, akkaynak2018revised} is governed by:
\begin{equation}
    \textbf{X}_{img}(x)=\textbf{Y}(x)\textbf{t}_{m}(x)+B(1-\textbf{t}_{m}(x))
    \label{eqn:IFM}
\end{equation}
where \(\textbf{X}_{img}(x)\) represents observed intensity at pixel \(x\), \(\textbf{Y}(x)\) corresponds to scene radiance or the true intensity of the object at pixel \(x\), which we aim to recover, \(\textbf{t}_{m}(x)\) denotes transmission map, referring to the fraction of light that successfully reaches the optical sensor from the object at pixel \(x\) without being scattered, \(B\) is the background light or ambient light, which is the light scattered into the camera due to the medium. The pixel \(x\in \left\{ 1,2,...,H\times W \right\}\), where \(H\) and \(W\) indicate the image’s height and width. Equation (\ref{eqn:IFM}) is rewritten to estimate scene radiance \(\textbf{Y}(x)\) by substituting \(\textbf{K}(x)=\frac{1}{\textbf{t}_{m}(x)}-1\).

\begin{equation}
    \textbf{Y}(x)=\textbf{K}(x)\textbf{X}_{img}(x)-\textbf{K}(x)\textbf{B}+\textbf{X}_{img}(x)
    \label{eqn:J(x)_modified}
\end{equation}

The estimates of \(\textbf{K}\in \mathcal{R}^{T \times 3 \times H \times W}\), and background light \(\textbf{B}\in \mathcal{R}^{T \times 3 \times H \times W}\) are subsequently passed into soft image reconstruction block together with the the hazy time-dependent input image sequence \(\textbf{X}_{img} \in \mathcal{R}^{T \times 3 \times H \times W}\).  The soft image reconstruction block is governed by (\ref{eqn:J(x)_modified}) and uses the inputs to predict the dehazed image \(\widehat{\textbf{Y}}\).

\subsection{Loss Function}

Underwater image dehazing presents unique challenges due to non-uniform light attenuation, scattering effects, and color distortions, which degrade visibility and structural integrity. Traditional loss functions such as MSE alone often fail to preserve fine details, contrast, and perceptual quality, as they primarily focus on pixel-wise intensity differences. To address these limitations, this research designs an application-specific loss function that integrates perceptual and structural quality metrics alongside conventional pixel-wise losses.

The proposed composite loss function ensures that the model effectively reduces haze while maintaining structural fidelity, preserving natural color distribution, and suppressing noise artifacts. The loss formulation combines MSE loss \cite{zhao2016loss}, SSIM loss \cite{zhao2016loss}, and Total Variance (TV) loss \cite{rudin1992nonlinear}, each contributing to different aspects of image enhancement:

Mean Squared Error (MSE) Loss: Minimizes direct pixel intensity differences to ensure reconstruction consistency.
\begin{equation}
    \centering
    \mathcal{L}_{MSE}(Y,\hat{Y})=\frac{1}{H\times W}\sum_{i=1}^{H}\sum_{j=1}^{W}\left ( Y_{i,j}-\hat{Y}_{i,j} \right )^{2}.
\end{equation}
Here, \(H\) and \(W\) denotes image's height and width, \(Y\) represents the reference image, \(\hat{Y}\) denotes the reconstructed image. \(\hat{Y}_{i,j}\) indicates the membrane potential (intensity) at position \((i,j)\) in the reconstructed image, and \(Y_{i,j}\) represents pixel intensity at position \((i,j)\) in the reference image.

Structural Similarity Index Measure (SSIM) Loss: Enforces structural and perceptual similarity to the reference image.
\begin{equation}
    \centering
    \mathcal{L}_{SSIM}(Y, \hat{Y}) = \frac{(2 \mu_Y \mu_{\hat{Y}} + C_1)(2 \sigma_{Y\hat{Y}} + C_2)}{(\mu_Y^2 + \mu_{\hat{Y}}^2 + C_1)(\sigma_Y^2 + \sigma_{\hat{Y}}^2 + C_2)}
\end{equation}
where \(\mu_{Y}\) and \(\mu_{\hat{Y}}\) are the means, \(\sigma _{Y}^{2}\) and \(\sigma _{\hat{Y}}^{2}\) are the variances, \(\sigma _{Y\hat{Y} }\) is the covariance. \(C_{1}\) and \(C_{2}\) are constants.

Total Variance (TV) Loss: Regularizes the dehazed output to maintain smoothness and suppress artifacts.

\begin{equation}
    \mathcal{L}_{TV}(\hat{Y}) = \frac{
        \left(
        \begin{aligned}
            &\sum_{c=1}^{C} \sum_{i=1}^{H-1} \sum_{j=1}^{W-1} \Big( \left( \hat{Y}_{c,i,j} - \hat{Y}_{c,i+1,j} \right)^2 \\
            &\quad + \left( \hat{Y}_{c,i,j} - \hat{Y}_{c,i,j+1} \right)^2 \Big)
        \end{aligned}
        \right)
    }{C \times H \times W}
\end{equation}

Net Loss: The cumulative loss function is formulated as the sum of weighted components, where the weighting factors are empirically determined to balance dehazing effectiveness, perceptual similarity, and spatial smoothness.

\begin{equation}
    \mathcal{L}_{net}=\mathcal{L}_{MSE}+ \alpha(1-\mathcal{L}_{SSIM})+\beta\mathcal{L}_{TV}.
    \label{eqn:loss_fn}
\end{equation}
where \(\alpha=0.5\) and \(\beta=0.25\) is selected with empirical analysis.

\section{Results and Analysis}

\subsection{Implementation Strategy}
The proposed snnTrans-DHZ architecture is implemented using SpikingJelly \cite{spikingjelly} and the PyTorch library. The model is trained on an NVIDIA A100 GPU equipped with 40GB of RAM. It undergoes 100 training epochs on the UIEB dataset and 50 epochs on the EUVP dataset. In both cases, the validation starts at the 20th epoch, allowing the model a warm-up period. To ensure optimal performance, the most effective model is selected by identifying the lowest validation loss recorded across epochs. The Adam optimizer \cite{kingma2014adam} is employed with a learning rate of 0.001 to ensure stable convergence and efficient gradient updates. All input images are scaled \(512 \times 512\) for both training and inference. The timestep \(T\) is selected as \(10\). The snnTrans-DHZ model is trained using the surrogate gradient-based BPTT strategy (refer to section \ref{sec:BPTT}) to update the network weights, ensuring \(\mathcal{L}_{net}\) minimization.

\begin{table*}[h!]
\caption{Performance evaluation of snnTrans-DHZ framework compared with UIE-SNN. The evaluation is based on algorithmic and hardware-oriented performance metrics. Both frameworks are trained and tested on UIEB and EUVP datasets.}
\label{table:c4_performance_eval}
\setlength{\tabcolsep}{3pt} 
\renewcommand{\arraystretch}{1.3} 
\centering
\resizebox{\textwidth}{!}{%
\begin{tabular}{c|c|c|c|c|c|c|c}
\hline
\multirow{3}{*}{\textbf{Dataset}} & \multirow{3}{*}{\textbf{Architecture}} & \multicolumn{2}{c|}{\textbf{Algorithmic Performance}} & \multicolumn{4}{c}{\textbf{Hardware-Oriented Metrics}} \\
\cline{3-8} 
& & \textbf{PSNR (dB)} & \textbf{SSIM} & \textbf{\# Parameters} & \textbf{SOPs} & \textbf{Energy} & \textbf{Energy Ratio} \\
& & ($\uparrow$) & ($\uparrow$) & \textbf{(M)} ($\downarrow$) & \textbf{(G)} ($\downarrow$) & \textbf{(J)} ($\downarrow$) & ($\times$) ($\downarrow$) \\
\hline \hline
\multirow{2}{*}{UIEB} & \textbf{snnTrans-DHZ} & \textbf{21.6773} & \textbf{0.8795} & \textbf{0.57} & \textbf{7.42} & \textbf{0.0151} & $1\times$ \\
& UIE-SNN & 17.7801 & 0.7454 & 31.03 & 147.49 & 0.1327 & $8.79\times$ \\
\hline
\multirow{2}{*}{EUVP} & \textbf{snnTrans-DHZ} & \textbf{23.4562} & \textbf{0.8439} & \textbf{0.57} & \textbf{8.43} & \textbf{0.0159} & $1\times$ \\
& UIE-SNN  & 23.1725 & 0.7890 & 31.03 & 84.65 & 0.0732 & $4.60\times$ \\
\hline
\end{tabular}}
\end{table*}

The snnTrans-DHZ model exhibits superior algorithmic performance, achieving higher PSNR and SSIM  across UIEB and EUVP datasets. The hybrid RGB-LAB color space and spiking transformer architecture used in snnTrans-DHZ enhanced the feature extraction and color consistency. This is critical for underwater visual tasks where color distortions and light scattering significantly degrade image quality. The incorporation of the custom loss function \(\mathcal{L}_{net}\) further optimizes perceptual quality by balancing pixel-wise prediction, structural similarity, and smoothness.

\subsection{Performance Evaluation}

A key advantage of snnTrans-DHZ is its drastically lower computational overhead. It requires only 0.57~M parameters compared to 31.03~M in UIE-SNN, making it significantly lighter and more efficient. The reduction in parameters directly impacts the computational complexity, allowing deployment on hardware with limited memory and processing capabilities, such as edge devices in underwater robotic platforms. Furthermore, snnTrans-DHZ significantly reduces the number of synaptic operations, requiring only 7.42~G and 8.43~G SOPs for UIEB and EUVP datasets, respectively, compared to 147.49~G and 84.65~G for UIE-SNN.

The energy consumption of snnTrans-DHZ (0.0151~J for UIEB, and 0.0159~J for EUVP) is significantly lower than UIE-SNN, which consumes 0.1327~J and 0.0732~J respectively. The energy ratio metric further highlights the disparity, showing that snnTrans-DHZ achieves the same task with 8.79\(\times\) and 4.60\(\times\) less energy consumption than UIE-SNN, making it ideal for low-powered underwater robots.

\begin{figure*}[t!]
    \centering
    \includegraphics[width=\textwidth]{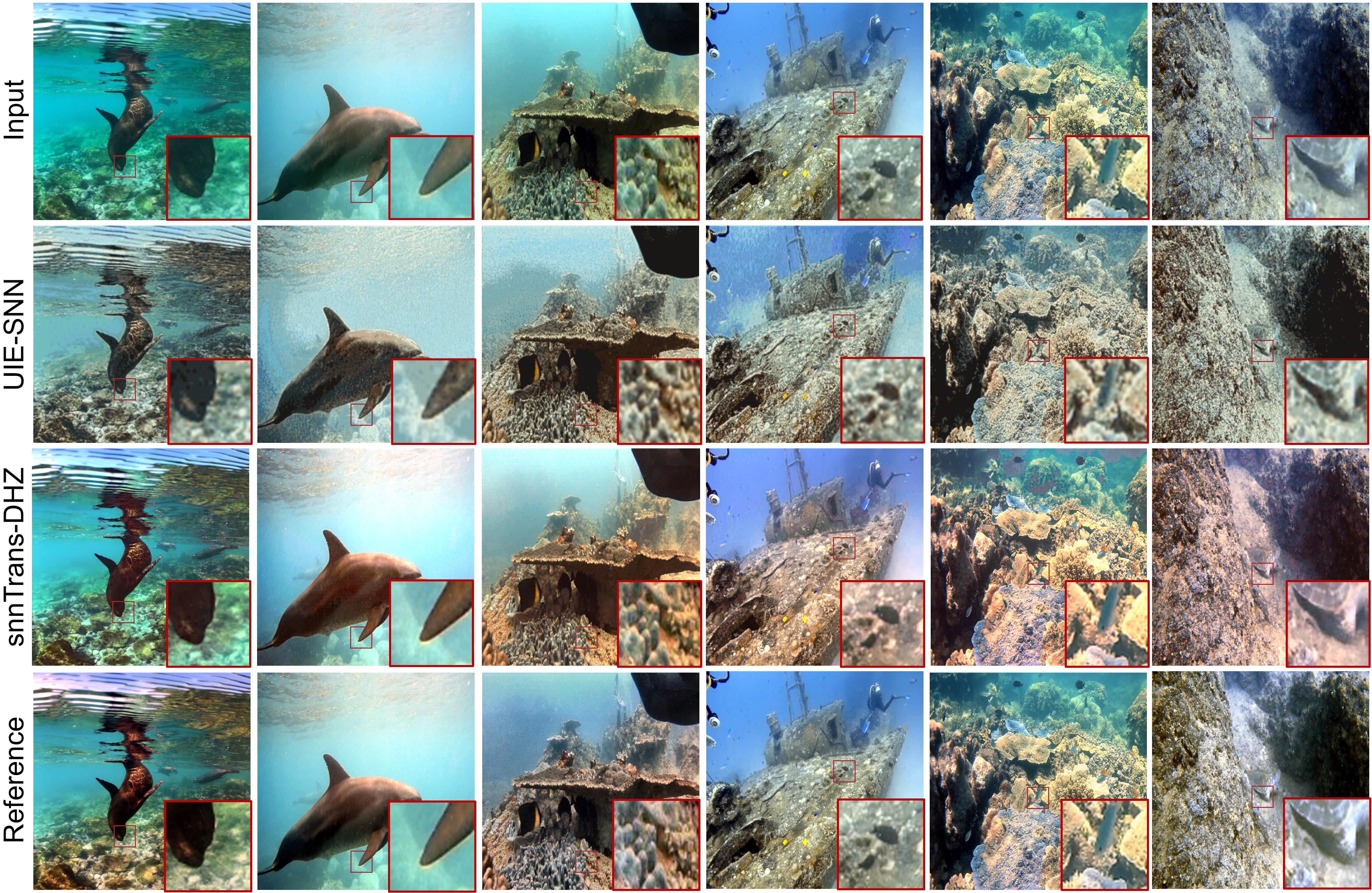}
    \caption{Qualitative analysis of the proposed snnTrans-DHZ framework on six different images from UIEB test samples. The top row indicates the unprocessed, raw underwater input images. The network-predicted images using the UIE-SNN and snnTrans-DHZ frameworks are shown in the second and third rows respectively. The last row represents the visibility-enhanced reference images.}
    \label{fig:c4_uieb_eval}
\end{figure*}

The qualitative analysis of the proposed snnTrans-DHZ framework is evaluated on UIEB, the real-world underwater dataset, and is presented in Fig. \ref{fig:c4_uieb_eval}. The qualitative comparison between snnTrans-DHZ and UIE-SNN highlights significant differences in image enhancement quality, particularly in terms of color restoration, structural detail preservation, and noise reduction. The visual results presented in the figure demonstrate that snnTrans-DHZ produces clearer, more vibrant, and artifact-free underwater images, making it a superior choice for real-world applications.

One of the most prominent improvements in snnTrans-DHZ is its ability to restore natural colors more effectively. Underwater images frequently exhibit a noticeable green or blue tint caused by light scattering and absorption. The images processed by UIE-SNN still retain a noticeable color cast, leading to duller and less realistic appearances. In contrast, snnTrans-DHZ successfully corrects color distortions, producing images with more natural and balanced tones. This improvement is particularly beneficial for marine biology and underwater exploration applications, where accurate color representation is crucial for species identification and habitat assessment.

\begin{table*}[t!]
\caption{Evaluation of snnTrans-DHZ under different ablation settings}
\label{table:c4_abla_alif_fx_gb}
\setlength{\tabcolsep}{3pt} 
\renewcommand{\arraystretch}{1.1} 
\centering
\resizebox{0.9\textwidth}{!}{%
\begin{tabular}{c|c|c|c|c|c|c}
\hline
\multicolumn{4}{c|}{\textbf{Ablation Settings}} & \textbf{PSNR} & \textbf{SSIM} & \textbf{\# Parameters} \\
\cline{1-4}
\multicolumn{2}{c|}{\textbf{Threshold}} & \multicolumn{2}{c|}{\textbf{Color Space}} & \textbf{(dB)} & & \textbf{(M)}\\
\cline{1-4}
\textbf{Fixed (0.5V)} & \textbf{Adaptive} & \textbf{RGB-LAB} & \textbf{RGB-Alone} & (\(\uparrow\)) & (\(\uparrow\)) & (\(\uparrow\))\\
\hline \hline
 & \checkmark & \checkmark & & \textbf{21.6773} & \textbf{0.8795} & 0.57 \\
\checkmark & & \checkmark & & 20.8279 & 0.8722 & 0.57\\
 & \checkmark & & \checkmark & 18.2622 & 0.6764 & \textbf{0.29} \\
\hline
\end{tabular}}
\end{table*}

Beyond color restoration, structural detail preservation is another key aspect where snnTrans-DHZ excels. The zoomed-in insets in the figure reveal that UIE-SNN struggles to retain fine textures, often introducing blurriness and loss of detail in complex regions such as fish skin, coral structures, and shipwrecks. On the other hand, snnTrans-DHZ enhances edge sharpness and texture clarity, making objects more distinguishable. This improvement is essential for underwater archaeology and object recognition, where fine-grained details play a crucial role in analysis.

Noise reduction and artifact suppression are additional strengths of the snnTrans-DHZ framework. Underwater images often contain artifacts due to poor visibility and environmental disturbances. The results indicate that UIE-SNN generates excessive noise, particularly in darker and smooth-surfaced regions, creating unnatural textures. In contrast, snnTrans-DHZ effectively reduces noise while maintaining sharpness, leading to more visually appealing and realistic outputs. Additionally, snnTrans-DHZ provides superior clarity in complex underwater scenes, such as shipwrecks and coral reefs. The method accurately reconstructs fine details, preserving object boundaries and intricate textures. This is particularly valuable for environmental monitoring, where capturing fine structural details is necessary for assessing coral health and detecting underwater debris.

\subsubsection{Effectiveness of snnTrans-DHZ on underwater robotic missions}

\begin{figure*}[h!]
    \centering
    \includegraphics[width=\textwidth]{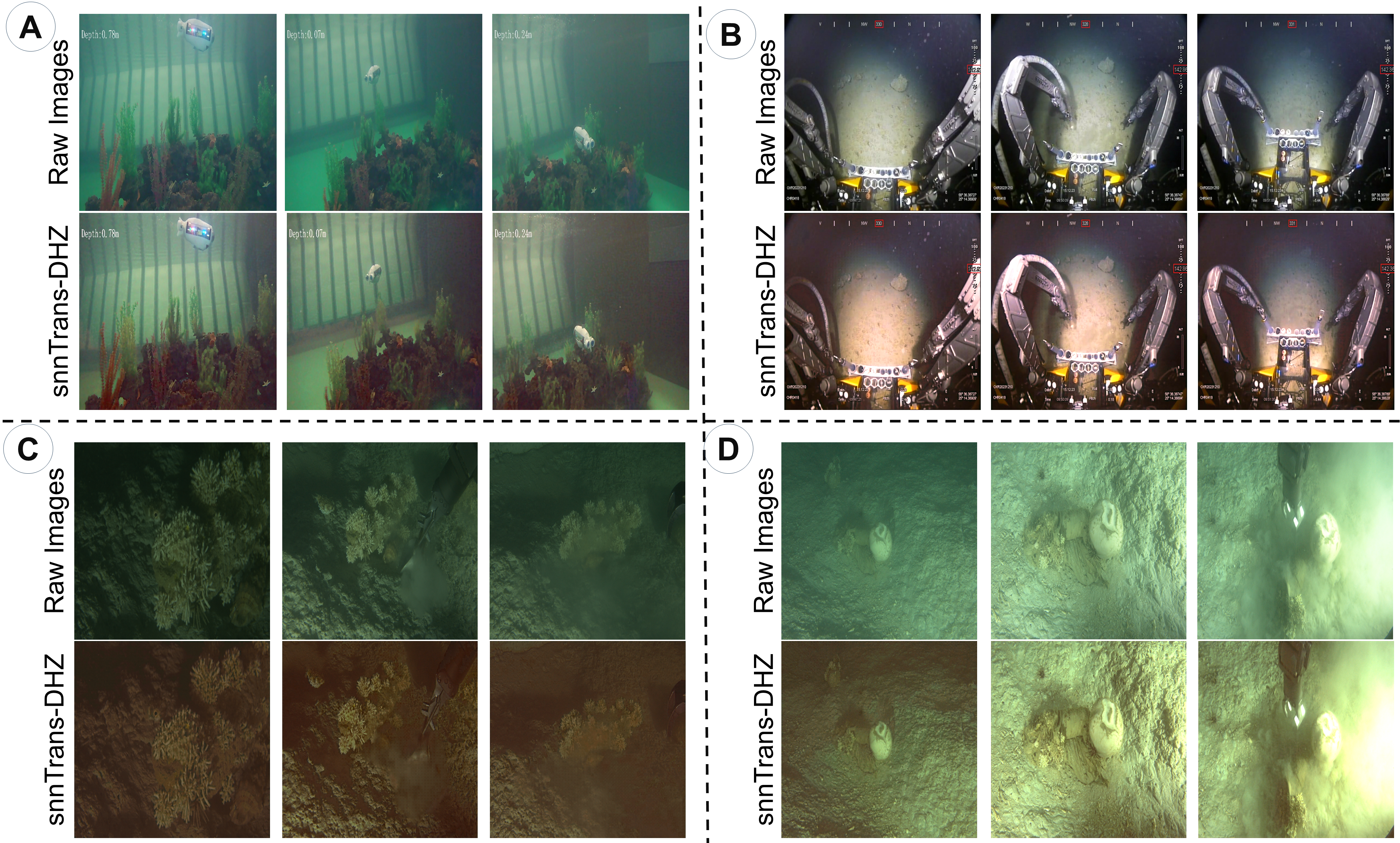}
    \caption{Qualitative analysis of snnTrans-DHZ on raw underwater images captured during various underwater missions. (A) The scenes are collected from the in-house Khalifa University Marine Robotics Pool Facility; (B)-(D) The underwater scenes collected during a 20-day expedition mission in the Arabian Gulf to explore the impacts of climate change on UAE waters (a collaboration between OceanX, M42 Healthcare, and Khalifa University).}
    \label{fig:c4_real}
\end{figure*}

Figure.\ref{fig:c4_real} presents a qualitative analysis of snnTrans-DHZ applied to raw underwater images captured during various underwater missions, showcasing its effectiveness in enhancing underwater visual perception. The results are divided into four sections: (A) indoor pool experiments conducted at Khalifa University Marine Robotics Pool Facility, and (B)-(D) real-world underwater scenes collected during a 20-day expedition in the Arabian Gulf, in collaboration with OceanX, M42 Healthcare, and Khalifa University.

In scenario (A), images from the controlled pool environment exhibit typical underwater distortions, including greenish color casts, reduced contrast, and blurriness. The application of snnTrans-DHZ successfully enhances these images by improving color balance and contrast, making objects such as corals and robotic vehicles more visually distinguishable. This demonstrates the model’s capability to function in structured testing environments, where controlled assessments of underwater imaging algorithms are crucial for real-world deployment.

The images in scenario (B) depict deep-sea intervention operations (at around 143~m depth) involving ROVs equipped with manipulators. The raw images suffer from poor lighting conditions, significant backscatter, and low visibility. The snnTrans-DHZ enhancement results show improved clarity, edge definition, and contrast, making fine details of the manipulator arms and seabed textures more visible. This enhancement is particularly beneficial for underwater inspection and intervention tasks, where operators rely on clear visual feedback to precisely manipulate objects.

The scenarios (C) and (D) capture marine ecosystems, particularly coral structures, and benthic organisms, affected by challenging underwater lighting conditions. The raw images in these scenes display severe color distortion and reduced visibility due to light absorption and scattering. The snnTrans-DHZ method enhances color fidelity, structural visibility, and scene interpretability, making marine biodiversity more distinguishable. Such improvements are critical for marine research and ecological monitoring, enabling more accurate assessments of coral health, habitat changes, and climate change effects on marine ecosystems.

The qualitative analyses presented in Fig. \ref{fig:c4_uieb_eval} and Fig. \ref{fig:c4_real} highlight the generalization capabilities of the proposed snnTrans-DHZ framework across diverse underwater environments. While the model was trained solely on the publicly available UIEB dataset (Fig. \ref{fig:c4_uieb_eval}), qualitative assessments on images from significantly different real-world underwater scenarios, including controlled pool facilities (Fig. \ref{fig:c4_real}A) and challenging deep-sea conditions (depths exceeding \(140~m\), Fig. \ref{fig:c4_real}B–D), demonstrate its effectiveness. The snnTrans-DHZ maintains robust dehazing performance and perceptual clarity across varying turbidity levels and distinct lighting conditions, underscoring its practical applicability and generalization beyond the training dataset.

Overall, the qualitative results demonstrate that snnTrans-DHZ effectively enhances underwater images across diverse conditions, ranging from structured test environments to real-world deep-sea and ecological exploration missions. The proposed method significantly improves contrast, color correction, and structural clarity, making it a valuable tool for underwater robotics, marine research, and environmental monitoring.

\begin{table*}[ht!]
\caption{Comparison of the snnTrans-DHZ framework with SOTA UIE methods using UIEB.}
\label{table:c4_sota_snnTrans_learn}
\begin{threeparttable}
\renewcommand{\arraystretch}{1.3} 
\adjustbox{width=\textwidth}{
\begin{tabular}{c|c|c|c|c|c|c}
\hline
 & & & &
\textbf{Number} & 
\textbf{Energy} & 
\textbf{Energy} \\

\textbf{Method} & \textbf{PSNR} & \textbf{SSIM} &  \(\#\)\textbf{Parameters} &
\textbf{of} & 
\textbf{($\downarrow$)} & 
\textbf{Ratio ($\downarrow$)$^{\#}$} \\
\hline \hline

\textbf{CNN-based method:} \hfill{} &  &  &  &  &  &  \\
WaterNet \hfill{}(2020) \cite{li2020underwater}  & 21.1700 & 0.8689 & 1.09 & 285.80 & 1.3147 &  87.07\(\times\) \\
Shallow-UWNet \hfill{}(2021) \cite{naik2021shallow}  & 18.2780 & 0.8550 & 0.22 & 216.14 & 0.9942 & 65.84\(\times\)\\
Ucolor \hfill{}(2021) \cite{li2021underwater}  & 18.8400 & 0.7612 & 157.42 & 887.80 & 4.0839 & 270.46\(\times\)\\
PUIE-Net \hfill{}(2022) \cite{fu2022uncertainty}  & 21.3800 & 0.882 & 1.40 &300.04 & 1.3802 & 91.40\(\times\)\\
PDCFNet  \hfill{}(2024) \cite{zhang2024pdcfnet} & \textbf{27.3700} & 0.9200 & 1.82 &172.77 & 0.7947 & 52.63\(\times\)\\
UIEConv \hfill{}(2024) \cite{du2024physical} & 24.1197 & 0.9283  & 3.31 &243.06 & 1.1181 & 74.05\(\times\)\\

\textbf{GAN-based method:} \hfill{} &  & &  &  &  &  \\
PUGAN \hfill{}(2023) \cite{cong2023pugan} & 19.9700 & 0.7848  & 95.66 &144.18 & 0.6632 & 43.92\(\times\)\\
LFT-DGAN \hfill{}(2024) \cite{zheng2024learnable} & 24.4591 & 0.8284 & 3.30 & 264.28 & 1.2157 & 80.51\(\times\)\\

\textbf{Attention-based method:} \hfill{} &  &  & &  &  &  \\
LANet \hfill{}(2022) \cite{liu2022adaptive}  & 24.0600  & 0.9085  & 5.15 & 355.37 & 1.6347 & 108.26\(\times\)\\
Deep-WaveNet \hfill{}(2023) \cite{sharma2023wavelength}  & 23.3000 & 0.9199 & - & 145.18 & 0.6678 & 44.22\(\times\)\\
DICAM \hfill{}(2024) \cite{tolie2024dicam} & 24.4300 & \textbf{0.9375} & - & 106.26 & 0.4888 & 32.37\(\times\)\\  

\textbf{Transformer-based method:} \hfill{} &  &  &  &  &  &  \\
U-Trans \hfill{}(2023) \cite{peng2023u} & 22.9100 & 0.9100 & 65.60 &132.04 & 0.6074 & 40.22\(\times\)\\
MFMN  \hfill{}(2024) \cite{zheng2024multi} & 24.2067 & 0.8992 & \textbf{0.20} &205.61 & 0.9458 & 62.64\(\times\)\\
MMIETransformer \hfill{}(2024) \cite{kulkarni2024multi} & 23.2800 & 0.9100 & 16.80 & 164.00 & 0.7544 & 49.96\(\times\)\\
WPFNet \hfill (2024) \cite{liu2024wavelet} & 22.3400 & 0.9072 & 2.16 & 40.20 & 0.1849 & 12.24\(\times\) \\
DDformer \hfill (2024) \cite{gao2024ddformer} & 23.5007 & 0.8944 & 7.58 & 35.74 & 0.1644 & 10.89\(\times\)\\
Ghost-UNet \hfill (2024) \cite{sun2024ghost} & 23.1600 & 0.8300 & 1.68 & 10.72 & 0.0493 & 3.26\(\times\)\\
Dynamic SpectraFormer \hfill (2024) \cite{hu2024dynamic} & 25.9800 & 0.9341 & 64.20 & 100.40 & 0.4618 & 30.58\(\times\)\\
CFPNet \hfill{}(2025) \cite{fu2024cfpnet} & 25.6900 & 0.8860 & 47.84 & 111.84 & 0.5145 & 34.07\(\times\)\\
\hline
\textbf{SNN-based method:} \hfill{} &  &  &  &  &  &  \\
UIE-SNN \hfill (2025) {\cite{sudevan2025underwater}} & 17.7801 & 0.7454 & 31.03 & 147.49 & 0.1327 & 8.79\(\times\) \\
\textbf{snnTrans-DHZ \hfill{} {[Ours]}} & 21.6773 & 0.8795 & 0.57 & \textbf{7.4156} & \textbf{0.0151} & \textbf{1\(\times\)} \\
\hline
\end{tabular}
}
\begin{tablenotes}
    \item \(\star\) Number of operations are measured in GSOPs for SNN-based method and in GFLOPs for others.
    \item \(\# \) The energy ratio is calculated as \(\frac{E_{method}}{E_{snnTrans-DHZ}}\).
\end{tablenotes}
\end{threeparttable}
\end{table*}

\begin{table*}[h!]
\caption{Performance evaluation of the snnTrans-DHZ framework on different image resolutions.}
\label{table:c4_abla_resolution}
\setlength{\tabcolsep}{3pt} 
\renewcommand{\arraystretch}{1.1} 
\centering
\resizebox{0.6\textwidth}{!}{%
\begin{tabular}{c|c|c|c|c}
\hline
\textbf{Image} & \textbf{PSNR} & \textbf{SSIM} & \textbf{SOPs} & \textbf{Energy} \\
\textbf{Resolution} & \textbf{(\textbf{dB})} (\(\uparrow\)) & (\(\uparrow\)) & \textbf{(G)} (\(\downarrow\)) & \textbf{(J)} (\(\downarrow\)) \\
\hline \hline
\(256 \times 256\) & 21.6988 & 0.8818 & 1.6711 & 0.0017 \\
\(512 \times 512\) & 21.6773 & 0.8795 & 7.4156 & 0.0150 \\
\(1024 \times 1024\) & 21.6587 & 0.8804 & 25.8353 & 0.0266\\
\hline
\end{tabular}}
\end{table*}

\subsection{Ablation Studies}
\subsubsection{Threshold Membrane Potential}
As shown in Table \ref{table:c4_abla_alif_fx_gb}, the model incorporating adaptive LIF neuron with learnable threshold membrane potential with RGB-LAB processing achieves the highest PSNR (\(21.6773~dB\)) and SSIM value \(0.8795\). This settings demonstrates that integrating ALIF neurons with hybrid color space effectively enhances model performance. These results highlight that adaptive neuron dynamics and combined luminance-chromaticity processing significantly contribute to better haze removal and perceptual image quality, without substantially increasing model complexity. 

Replacing adaptive ALIF neurons with fixed-threshold neurons while retaining RGB-LAB hybrid processing leads to a noticeable performance drop, with PSNR and SSIM decreasing to \(20.8279~dB\) and, \(0.8722\) respectively. This reduction indicates the pivotal role played by the adaptive threshold in ALIF neurons, emphasizing their contribution to modeling temporal dynamics and improving the overall quality of underwater image reconstruction. The consistent number of parameters confirms the performance variation arises solely from adaptive neuron mechanisms rather than model complexity.

\subsubsection{Effect of Hybrid Color Space Processing}
It is evident from Table \ref{table:c4_abla_alif_fx_gb} that processing images with a hybrid RGB-LAB color space substantially improves the dehazing performance of the model. The results clearly demonstrates the critical advantage provided by hybrid RGB-LAB processing, suggesting that separately handling luminance and chromaticity channels greatly enhances dehazing and image visibility. The reduced parameter count (\(0.29~M\)) here is due to the omission of the LAB processing branch, underscoring that the observed performance drop originates primarily from the absence of hybrid color space processing rather than the adaptive ALIF mechanism.

\subsubsection{Image Resolution}
To evaluate the efficiency and scalability of snnTrans-DHZ for underwater robotics applications, we analyze its performance across different input image resolutions. The results, summarized in Table \ref{table:c4_abla_resolution}, provide insights into the trade-offs between image quality, computational complexity, and energy consumption at varying resolutions. 

The PSNR and SSIM values remain relatively stable across resolutions, indicating that snnTrans-DHZ maintains high-quality image enhancement regardless of input size. The PSNR fluctuates slightly around \(21.68~dB\), and the SSIM consistently exceeds \(0.88\), demonstrating the model’s robustness in restoring underwater images with minimal degradation, even at lower resolutions. This is crucial for underwater robotics, where efficient processing of lower-resolution images can significantly reduce latency without compromising perceptual quality.

However, increasing resolution leads to a sharp rise in computational cost. The SOPs grow from \(1.6711~G\) at \(256\times 256\) image resolution to \(25.8353~G\) at \(1024 \times 1024\) image resolution, demonstrating a nearly \(15.46\times\) increased number of synaptic operations. Similarly, energy consumption rises from \(0.0017~J\) to \(0.0266~J\), highlighting the increased computational burden of higher-resolution inputs. While lower-resolution image processing is often preferred for real-time underwater robotics applications due to its efficiency, there are critical scenarios where high-resolution image processing becomes essential. Tasks such as marine biodiversity monitoring, archaeological site mapping, deep-sea exploration, and defect detection in underwater infrastructure require fine-grained visual details to ensure accurate interpretation and analysis. In such cases, the ability to process high-resolution images efficiently is a key advantage.

For practical deployment in underwater robots, \(512 \times 512\) resolution emerges as the optimal choice, offering a strong balance between quality (PSNR=\(21.6773~dB\), SSIM=\(0.8795\)) and efficiency (SOPs=\(7.4156~G\), energy=\(0.0150~J\)). This resolution provides sufficient details for object detection and navigation while maintaining a manageable computational load, facilitating real-time processing in energy-constrained environments.

These findings highlight the scalability of snnTrans-DHZ, making it adaptable for a range of underwater robotics applications. Depending on mission requirements, lower resolutions can be leveraged for fast, low-power processing, while higher resolutions may be used selectively when detailed scene understanding is necessary. 

\section{Discussions}

The comparison between the proposed snnTrans-DHZ algorithm and SOTA learning-based UIE methods is presented in Table \ref{table:c4_sota_snnTrans_learn}. The comparison of snnTrans-DHZ with SOTA UIE methods highlights its balance between performance and efficiency. While snnTrans-DHZ yields PSNR score of \(21.6773~dB\) and SSIM as \(0.8795\), which is lower than leading transformer-based methods like Dynamic SpectraFormer (PSNR of \(25.98~dB\), SSIM of \(0.9341\)) and CFPNet (PSNR of \(25.69~dB\), SSIM of \(0.8860\)). The proposed snnTrans-DHZ framework  significantly outperforms other compact models in computational efficiency. The higher image quality from transformer-based models comes at the cost of significantly higher computational complexity and energy consumption, making them less suitable for resource-constrained environments.

A key advantage of snnTrans-DHZ is its lightweight design, with only \(0.57M\) parameters, making it one of the most compact models in the comparison. In contrast, Dynamic SpectraFormer and CFPNet require \(64.2M\) and \(47.84M\) parameters, respectively, making them 112× and 84× larger. Even the lightweight MFMN model, which has a smaller parameter count (\(0.20M\)), still consumes significantly higher energy. This compact architecture of SNNTrans-DHZ enables efficient deployment on edge devices without the memory constraints of larger transformer-based networks.

Another critical metric is computational cost. snnTrans-DHZ requires only \(7.4156G\) SOPs, whereas Dynamic SpectraFormer (\(100.4G\) FLOPs) and CFPNet (\(111.84G\) FLOPs) are \(13.54\times\) and \(15.08\times\) more computationally intensive, respectively. Even relatively efficient transformer-based methods like DDformer (\(35.74G\) FLOPs) and Ghost-UNet (\(10.72G\) FLOPs) remain at least \(4.82\times\) and \(1.44\times\) more computationally demanding than snnTrans-DHZ. This drastic reduction in computational requirements allows snnTrans-DHZ to operate with notably lower latency, making it well-optimized for real-time underwater robotic applications.

Energy efficiency is another defining strength of SNNTrans-DHZ. It consumes only \(0.0151 ~J\), which is substantially lower than all other SOTA UIE methods. Ghost-UNet, the most energy-efficient transformer-based model, requires \(0.0493~J\), which is \(3.26\times\) more than snnTrans-DHZ. Dynamic SpectraFormer (\(0.4618~J\)) and CFPNet (\(0.5145~J\)) require \(30.58\times\) and \(34.07\times\) more energy, respectively. This significant energy savings extends the operational lifespan of battery-powered underwater robotics, making snnTrans-DHZ an optimal choice for long-duration missions.

Some CNN-based models, such as PDCFNet (\(27.37~dB\), \(0.9200\) SSIM) and UIEConv (\(24.12~dB\),\( 0.9283\) SSIM), achieve higher PSNR and SSIM values. However, they still exhibit considerably higher energy consumption and computational costs. For example, PDCFNet consumes \(0.7947~J\) (\(52.63\times\) more than snnTrans-DHZ, while UIEConv requires \(1.1181~J\) (\(74.05\times\) more). Thus, snnTrans-DHZ offers a compelling alternative that balances image quality, efficiency, and deployability.

In summary, while snnTrans-DHZ does not achieve the absolute highest image enhancement quality, it significantly outperforms all competing models in both computational efficiency and energy savings. Its lightweight architecture, low power consumption, and ability to perform real-time processing make it highly practical for underwater robotic applications, where resource constraints are a critical factor. This substantial energy savings makes snnTrans-DHZ a more practical solution for battery-powered underwater robotics, extending mission durations while maintaining reliable enhancement quality.

\section{Conclusions and Future Research Directions}
This article detailed the development of snnTrans-DHZ, a lightweight and energy-efficient spiking transformer-based framework for underwater image dehazing. The snnTrans-DHZ algorithm achieves PSNR = \(21.6773~\text{dB}\) and SSIM = \(0.8795\) on UIEB, and PSNR = \(23.4562~\text{dB}\) and SSIM = \(0.8439\) on EUVP. The total network parameters in the snnTrans-DHZ architecture are \textbf{\(0.5670~M\)}. Compared to the existing SOTA image enhancement methods, the snnTrans-DHZ architecture could significantly reduce the model complexity and energy consumption. Future research will focus on deploying the snnTrans-DHZ algorithm onto an edge device and analyzing its applicability in real-world deployments. Since the number of parameters is less than one million, it can be deployed on the Loihi neuromorphic chip to evaluate its real-time performance on neuromorphic hardware.
\ifCLASSOPTIONcaptionsoff
  \newpage
\fi

\bibliographystyle{ieeetr}
\bibliography{snnTrans}

\end{document}